\documentclass[sigconf]{acmart}
\AtBeginDocument{%
  }

\copyrightyear{2025}
\acmYear{2025}
\setcopyright{cc}
\setcctype{by}
\acmConference[KDD '25]{Proceedings of the 31st ACM SIGKDD Conference on Knowledge Discovery and Data Mining V.2}{August 3--7, 2025}{Toronto, ON, Canada}
\acmBooktitle{Proceedings of the 31st ACM SIGKDD Conference on Knowledge Discovery and Data Mining V.2 (KDD '25), August 3--7, 2025, Toronto, ON, Canada}
\acmDOI{10.1145/3711896.3711896.3737376}
\acmISBN{979-8-4007-1454-2/2025/08}
\usepackage{amsmath}
\usepackage{graphicx}
\usepackage{algorithm}
\usepackage{algorithmic}
\usepackage{hyperref}
\usepackage{booktabs, tablefootnote}  
\usepackage{multirow}  
\usepackage{graphicx}  
\usepackage{subcaption}
\usepackage{lipsum} 
\usepackage{bbm}
\usepackage{enumitem,multirow,graphicx,subcaption,multicol,lipsum,float,adjustbox}
\usepackage{threeparttable}
\usepackage{mathtools}
\usepackage{xcolor,  colortbl}
\usepackage{makecell}
\usepackage{dblfloatfix}
\usepackage{caption}

\usepackage{listings}  
\usepackage{array}     
\usepackage{adjustbox} 
\renewcommand{\arraystretch}{1.1} 

\newcommand{\smallsection}[1]{{\vspace{0.05in} \noindent \bf {#1\hspace{5pt}}}}
\begin{document}

\title{Delving into Instance-Dependent Label Noise in Graph Data:
A Comprehensive Study and Benchmark}


\author{Suyeon Kim}
\affiliation{%
  \institution{Pohang University of \\ 
  Science and Technology}
  \city{Pohang}
  \country{Republic of Korea}}
\email{kimsu@postech.ac.kr}

\author{SeongKu Kang}
\authornote{Co-corresponding authors}
\affiliation{%
  \institution{Korea University}
  \city{Seoul}
  \country{Republic of Korea}}
\email{seongkukang@korea.ac.kr}

\author{Dongwoo Kim}
\affiliation{%
  \institution{Pohang University of \\ 
  Science and Technology}
  \city{Pohang}
  \country{Republic of Korea}}
\email{dongwoo.Kim@postech.ac.kr}

\author{Jungseul Ok}
\affiliation{%
  \institution{Pohang University of \\ 
  Science and Technology}
  \city{Pohang}
  \country{Republic of Korea}}
\email{jungseul.ok@postech.ac.kr}

\author{Hwanjo Yu}
\authornotemark[1]
\affiliation{%
  \institution{Pohang University of \\ 
  Science and Technology}
  \city{Pohang}
  \country{Republic of Korea}}
\email{hwanjoyu@postech.ac.kr}

\renewcommand{\shortauthors}{Suyeon Kim, SeongKu Kang, Dongwoo Kim, Jungseul Ok, \& Hwanjo Yu.}

\begin{abstract}
Graph Neural Networks (GNNs) have achieved state-of-the-art performance in node classification tasks but struggle with label noise in real-world data. 
Existing studies on graph learning with label noise commonly rely on class-dependent label noise, overlooking the complexities of instance-dependent noise and falling short of capturing real-world corruption patterns. 
We introduce BeGIN (Benchmarking for Graphs with Instance-dependent Noise), a new benchmark that provides realistic graph datasets with various noise types and comprehensively evaluates noise-handling strategies across GNN architectures, noisy label detection, and noise-robust learning.
To simulate instance-dependent corruptions, BeGIN introduces algorithmic methods and LLM-based simulations.
Our experiments reveal the challenges of instance-dependent noise, particularly LLM-based corruption, and underscore the importance of node-specific parameterization to enhance GNN robustness. 
By comprehensively evaluating noise-handling strategies, BeGIN provides insights into their effectiveness, efficiency, and key performance factors. 
We expect that BeGIN will serve as a valuable resource for advancing research on label noise in graphs and fostering the development of robust GNN training methods.
The code is available at \url{https://github.com/kimsu55/BeGIN}. 
\end{abstract}



\received{24 February 2025}
\received[accepted]{16 May 2025}

\maketitle

\section{Introduction}
Graph Neural Networks (GNNs) have demonstrated remarkable performance in node classification tasks~\cite{maekawa2022beyond, rong2019dropedge, xiao2022graph}, but their effectiveness heavily depends on clean and accurately labeled datasets~\cite{dai2021nrgnn, yuan2023learning, qian2023robust}. 
Unfortunately, real-world graph data often contains mislabeled instances due to an unreliable data collection process~\cite{ji2023drugood}, human annotation errors~\cite{georgakopoulos2016weakly}, and adversarial attacks~\cite{zhang2020adversarial}, which negatively impact model robustness.
Moreover, since GNNs rely on a message-passing mechanism to aggregate information from neighboring nodes, label noise can propagate throughout the graph, posing significant challenges to model reliability and generalization.  
Consequently, learning with label noise (LLN) in graphs emerges as a critical research challenge.

While there have been attempts to apply LLN for graph data \cite{wang2024noisygl,cheng2024resurrecting, zhu2024robust, du2021noise, li2024contrastive}, they commonly rely on \textit{class-dependent} label noise, where corruption follows a fixed transition pattern based on class labels. 
However, real-world noise is often \textit{instance-dependent}~\cite{wei2021learning, menon2018learning, xia2020part}, where the corruption of individual nodes is influenced by various factors such as node features, graph topology, and contextual ambiguities~\cite{goldberger2017training}. 
Models trained on naive class-dependent noise fail to capture real-world corruption patterns, resulting in suboptimal performance and poor generalization to real-world applications.
As shown in computer vision, efforts to simulate and benchmark realistic noises have driven rapid advancements in noise-robust learning methods, evolving from simple uniform noise~\cite{van2015learning, ghosh2017robust, yu2023delving} to instance-dependent~\cite{xia2020part, jiang2020beyond} and human-annotated noisy datasets~\cite{peterson2019human, xiao2015learning, wei2021learning}. 
However, similar studies remain limited in graph learning, highlighting the need for more realistic noise simulations and the establishment of comprehensive benchmarks.

To address this gap, we present a new benchmark called \textbf{BeGIN} (\textbf{Be}nchmarking for \textbf{G}raph with \textbf{I}nstance-dependent \textbf{N}oise).
BeGIN provides \textbf{benchmark datasets} comprising $10$~graph datasets with six different noise types, along with \textbf{evaluation benchmark} for various noise-handling strategies, including diverse GNN architectures, noisy label detection, and noise robust learning.
To simulate more realistic noise beyond class-dependent assumptions, we first introduce different types of instance-dependent label noise.
Specifically, we incorporate both \textit{algorithmic noise injection}, which systematically perturbs labels based on graph topology, node features, and model confidence, and \textit{Large Language Model (LLM)-based noise simulation}, which mimics human annotation errors in scenarios where textual information heavily influences the labeling process. 
Building on the noise injection strategies above, we apply these instance-dependent corruptions to a variety of real-world graphs with diverse characteristics, including varying numbers of nodes, average degrees, and homophily ratios. 
These datasets also span multiple domains, such as citation networks, e-commerce, and web pages, ensuring a broad and realistic evaluation setting.

Through extensive experiments on the constructed benchmark datasets, we comprehensively analyze the impact of various label noise on GNN performance and identify key insights for developing noise-handling strategies.
Notably, instance-dependent noise, especially that generated by large language models (LLMs), poses substantial challenges due to its structured mislabeling and skewed corruption patterns, which reflect instance-level uncertainties rather than random perturbations. 
In contrast, class-dependent noise introduces corruption in a more uniform and class-agnostic manner, leading to less complex learning dynamics. 
Our findings also highlight that parameterizing the center node separately from its neighbors, as done in GraphSAGE, can significantly enhance robustness by reducing the propagation of errors during neighborhood aggregation.

Furthermore, we provide a comprehensive evaluation benchmark for methods addressing label noise.
Our study encompasses both \textit{noisy label detection} and \textit{noise robust learning}, two key strategies for handling corrupted data.
Specifically, to unveil an effective strategy for detecting noisy labels, we analyze the effectiveness of various approaches, including leveraging model loss trajectory and a supervised noisy label detector.
We also discuss the applicability of two distinct types of information, feature similarity and topology-based label consistency, to further enhance noisy label detection in graph data.
For noise-robust learning, we conduct an extensive comparison of existing methods and evaluate their effectiveness.
Additionally, we examine the training efficiency of various learning methods, providing a reliable guideline as a foundation for developing robust models to handle label noise in graph data.
By analyzing different noise-handling techniques, we offer insights into the robustness of existing methods and identify key factors influencing performance under label noise. 
We expect that this benchmark will serve as a valuable resource for advancing research in reliable graph learning.
Our key contributions are: 
\begin{itemize}[leftmargin=*] \vspace{-\topsep}

    \item We introduce BeGIN, which provides extensive benchmark datasets with various instance-dependent noise types and comprehensive evaluation benchmarks of different noise-handling strategies.

    \item We reveal the limitations of class-dependent noise and propose corruption processes that systematically inject instance-dependent noise using an algorithmic approach and LLM simulation.
    
    \item Our analyses provide a guideline for the development and evaluation of graph learning for label noise, encompassing GNN architecture, noisy label detection, and noise robust learning.
    
\end{itemize}


\section{Label Noise for Graph Data}
\label{sec:Label Noise for Graph Data}

\subsection{Preliminaries}

\noindent
\textbf{Notations.}
We define a graph $G$ as a combination of its topology and feature components, denoted as $G = (G^t, G^f)$. 
The topology component, $G^t = (V, E, \mathrm{X}^e)$, consists of a set of nodes $V = \{v_1, \ldots, v_N\}$, a set of edges $E \subseteq V \times V$, and an edge feature matrix $\mathrm{X}^e \in \mathbb{R}^{|E| \times d_e}$. 
The feature component, $G^f$, has a node feature matrix $\mathrm{X}^v \in \mathbb{R}^{N \times d_v}$, where each row $\mathbf{x}_i \in \mathbb{R}^{d_v}$ represents the feature vector of node $v_i$.

\smallsection{Problem formulation.}
In the $C$-class node classification task, we denote a dataset as $D=\{G, \mathrm{Y}\}$, where $\mathrm{Y}\in \mathbb{R}^{N}$ is the ground-truth label for nodes and $y_i\in\{1,..., C\}$ denotes the label of node $v_i$.
In this paper, we aim to construct a noisy dataset $\tilde{D}=\{G, \mathrm{Y}, \tilde{\mathrm{Y}}\}$, where $\tilde{\mathrm{Y}}$ represents the noisy label for nodes, and there exists $v_i\in V$ such that $y_{i} \neq \tilde{y}_{i}$. 
The noise rate $\eta$ is the fraction of nodes whose labels have been corrupted.

\smallsection{Transition probability.}
We define the transition probability as the likelihood that a node’s true label is flipped to a different class.
Transition probability for $N$ nodes in a dataset is denoted as $\mathrm{T}_D\in \mathbb{R}^{N\times C}$, where each row $\mathrm{T}_D(v_i)\in\mathbb{R}^{C}$ corresponds to the transition probability vector of node $v_i$, and $\mathrm{T}_D(v_i, c)$ denotes the probability of node $v_i$ being flipped to class $c$.   
Additionally, the transition matrix $\mathrm{T} \in \mathbb{R}^{C \times C}$ provides a summarized view of these probabilities at the class level.  
Each entry $\mathrm{T}_{i,j}$ represents the class-wise likelihood of a label corruption from class $i$ to class $j$.

\subsection{Class-dependent Label Noise} 
\label{Sec:class_dependent}
To the best our knowledge, all prior works on label noise in graphs~\cite{wang2024noisygl,cheng2024resurrecting, zhu2024robust, du2021noise, li2024contrastive} have focused on class-dependent noise, where the probability of mislabeling depends only on the true class label, i.e., $\mathrm{T}_{i,j}=\mathbb{P}(\tilde{Y} = j \mid Y = i)$. 
This approach assumes that corruption follows a fixed class-level transition pattern independent of individual node characteristics such as features or graph topology. 

Two widely used strategies are \textbf{uniform} noise and \textbf{pairwise} noise. 
Uniform noise occurs when the probability of label transitions is equally distributed across all alternative classes, while pairwise noise refers to preferential mislabeling between specific class pairs\footnote{We adhere to the standard practice of assigning labels to the subsequent class~\cite{wang2024noisygl}.}.
However, it often falls short of reflecting the complexities of real-world mislabeling, where instance-level\footnote{In this work, we use the terms `instance' and `node' interchangeably.} and contextual factors play a significant role.
Moreover, pairwise noise is practical only when the noise rate does not exceed 0.5.

\subsection{Algorithmic Approach for Instance-dependent Label Noise}
Real-world noise often exhibits instance-dependent characteristics~\cite{wei2021learning, berthon2021confidence}.
However, modeling transition probability for individual nodes is challenging because it can be influenced by various factors, including a node's label, position in the graph, features, and the labels of its neighboring nodes.
To address this, we introduce a new algorithmic approach for generating instance-dependent label noise for graphs.
It consists of two stages: first compute transition probabilities based on a noise modeling strategy and then apply label corruption according to the obtained probabilities.
We first explain three noise modeling strategies used in this work, followed by a detailed label corruption process.

\subsubsection{\textbf{Noise modeling strategies for graph data}}
\label{sec:noise_modeling}
To reflect the complexity of real-world noise, we introduce three distinct modeling strategies that leverage graph topology, input features, and predicted confidence.
Each strategy generates a unique transition probability based on its respective noise type: \textit{topology}, \textit{feature}, and \textit{ confidence} of a trained GNN model.

\smallsection{Topology-based noise.} 
This strategy is devised to reflect the impact of a graph's structural properties on label corruption: $\mathrm{T}_{i,j}=\mathbb{P}(\tilde{Y}=j \mid Y = i, G^t, Y)$. 
The core idea is that a node whose label is inconsistent with its neighbors is more susceptible to corruption, making it more likely to be mislabeled into the predominant class among its neighboring nodes.
A naive implementation would simply compute statistics of $h$-hop neighboring nodes.
However, this approach fails to fully capture the graph's topological structures.

To refine the selection of neighboring nodes, we use Personalized PageRank (PPR)~\cite{page1999pagerank}, which computes the relative importance of neighboring nodes across the graph by adjusting global and local significance.
It models an infinite random walk from a target node $v_i$ with a return probability $\alpha$ at each step.
Leveraging this concept, we quantify the influence of neighboring nodes using the PPR vector~\cite{gasteiger2019diffusion, hou2021massively}: $\pi^{\mathrm{ppr}}({v_i})=\alpha(\mathbf{I}-(1-\alpha)\mathbf{D}^{-1}\mathbf{A})^{-1}\mathbf{e}_{v_i}\in \mathbb{R}^N$, where $\mathbf{I}$ is the identity matrix, $\mathbf{D}$ is a diagonal degree matrix where the $i$-th diagonal entry corresponds to the out-degree of node $v_i$, $\mathbf{A}$ is the adjacency matrix, and $\mathbf{e}_{v_i}$ is one-hot vector with the $i$-th position set to $1$. 
We fix $\alpha$ to 0.9 throughout the experiments.
Based on this influence, we compute the relative weight of neighboring nodes belonging to class $c$, denoted as $q_{v_i}^{(c)}={\sum_{v_j \in V, y_j=c} \pi^{\mathrm{ppr}}(v_i, v_j)}/{\| \pi^{\mathrm{ppr}}({v_i}) \|_1}$, and subsequently derive the transition probability across all classes as $\mathrm{T}_D(v_i)=[ q_{v_i}^{(c)}]_{c=1}^{C}$.

\smallsection{Feature-based noise.} 
This strategy is devised to integrate node features into the transition probability: $\mathrm{T}_{i,j}=\mathbb{P}(\tilde{Y}=j \mid Y=i, G^f, Y)$.  
The key idea is that a node whose features deviate from its class representation is more susceptible to label corruption and prone to being mislabeled into a class with a similar feature pattern.
The class representation is computed as the mean feature value of its nodes: $\mathbf{z}_c = \frac{1}{|\{y_i = c \mid v_i \in V \}|} \sum_{v_i \in V, y_i = c} \mathbf{x}_i$.
The similarity between a node feature and the class representation is calculated using cosine similarity as: $s_c(v_i)=\operatorname{sim}_{\cos}(\mathbf{x}_i, \mathbf{z}_c)$. 
Finally, the transition probability is derived as $\mathrm{T}_D(v_i) = [\frac{s_c(v_i)}{\sum_{c'=1}^{C} s_{c'}(v_i)}]_{c=1}^{C}$.

\smallsection{Confidence-based noise.} 
This strategy is devised to consider both topological structures and node features using a GNN-based model: 
$\mathrm{T}_{i,j}=\mathbb{P}(\tilde{Y} = j \mid Y = i, f(G, Y))$, where $f(\cdot)$ denotes the model's predicted probability distribution across $C$ classes. 
The main idea is that a node with a low prediction probability for its given class is more susceptible to label corruption and prone to being mislabeled into a class with high probability.
We train a GNN~\cite{kipf2016semi} on dataset $D$ until it achieves the best validation accuracy. 
Refer to the Appendix~\ref{apx:prompt} for the detailed procedure.
The transition probability is denoted as $\mathrm{T}_D = f(G, \mathrm{Y})\in \mathbb{R}^{N\times C}$.

\vspace{0.02in}
\textbf{Remarks.} 
While confidence-based noise appears highly intuitive and is expected to be effective, we empirically found that it often fails to fully resemble real-world noise. We conjecture that this occurs for two possible reasons:
(1) The generated noise is model-dependent. 
This means that the noise patterns reflect the model's inductive bias, which does not necessarily align with real-world noise. 
(2) Neural networks are known to be highly vulnerable to the overconfidence problem~\cite{guo2017calibration}. 
Specifically, they often assign excessively high probabilities to a single class while assigning overly low probabilities to the remaining classes.
As a result, the model's output often fails to accurately capture correlations among the classes, potentially leading to suboptimal results.
For a detailed analysis, please refer to Section \ref{sec:LLM_challenge}.

\begin{algorithm}
\small
\caption{Instance-dependent label corruption process}
\label{alg:instance_corruption}
\begin{algorithmic}[1]
    \REQUIRE Transition probability $\mathrm{T}_D \in \mathbb{R}^{N \times C}$, noise rate $\eta$, labels $\mathrm{Y} \in \mathbb{R}^{N}$, number of classes $C$
    \ENSURE Noisy labels $\tilde{\mathrm{Y}}$

    \STATE Compute the number of corrupted instances: $N_c \gets \lfloor N \times \eta \rfloor$
    \STATE Compute corruption probability : $\mathrm{P}^{cor}(v_i) \gets 1 - \mathrm{T}_D[i, \mathrm{Y}_i], \quad \forall i \in \{1, \dots, N\}$
    \STATE Normalize: $\mathrm{P}^{cor} \gets \mathrm{P}^{cor} / \sum \mathrm{P}^{cor}$
    \STATE Sample $N_c$ indices for corruption: $I \gets \text{Sample}(\mathrm{Y}, N_c, \mathrm{P}^{cor})$
    \STATE Initialize noisy labels: $\tilde{\mathrm{Y}} \gets \mathrm{Y}$

    \FOR{each $i \in I$}
        \STATE Modify transition probability: $\mathrm{T}'_D[i, :] \gets \mathrm{T}_D[i, :]$
        \STATE Set $\mathrm{T}'_{D}[i, \mathrm{Y}_i] = 0$, then normalize $\mathrm{T}'_D[i, :]$
        \STATE Sample a new label: $\tilde{\mathrm{Y}}[i] \gets \text{Sample}(C, 1, \mathrm{T}'_D[i, :])$
    \ENDFOR

    \RETURN $\tilde{\mathrm{Y}}$
\end{algorithmic}
\end{algorithm}

\subsubsection{\textbf{Instance-dependent Corruption Process}}
Given transition probabilities $\mathrm{T}_D$ and the overall noise rate $\eta$, we construct a noisy dataset.
Algorithm \ref{alg:instance_corruption} describes the label corruption process.
First, we compute the \textit{corruption probability}, which represents the likelihood of each node being corrupted, as $\mathrm{P}^{cor}(v_i)= \sum_{c\in C \setminus {y_i}}\mathrm{T}_D(v_i, c)$.
We then sample a pre-defined number ($N \times \eta $) of nodes in proportion to their corruption probabilities. 
Finally, for each sampled node, we corrupt its label according to its transition probability.

\begin{table*}[t]
    \centering
    \caption{Dataset statistics and node classification accuracy (\%) of GraphSAGE under different noise types.
    Results for other GNN models are provided in Appendix \ref{apx:exp_robust}. Values represent the mean $\pm$ standard deviation over 10 independent runs.}
    \label{tab:dataset}
        \resizebox{0.99\textwidth}{!}{%
        \begin{threeparttable}
        \begin{tabular}{lccccccc|cccc}
        \toprule
         & &  \multicolumn{6}{c|}{Homophily datasets } & \multicolumn{4}{c}{Heterophily datasets\tnote{c} \enspace\cite{McCallum_4Universities}}  \\

        & &  Cora-ML \cite{bojchevski2017deep} & WikiCS \cite{mernyei2020wiki} & Product-s \cite{he2023harnessing} & Children \cite{ni2019justifying} & History \cite{ni2019justifying} & Photo \cite{shchur2018pitfalls} & Cornell & Texas & Washington & Wisconsin   \\
        \midrule
        
        \multirow{5}{*}{\rotatebox{90}{\textbf{Statistics}}}

        & \# Classes  & 7 & 10 & 44 & 24 & 12 & 12 & 5 & 5 & 5 & 5 \\
        & \# Nodes  & 2,995 & 11,701 & 54,025 & 76,875 & 41,551 & 48,362 & 191 & 187 & 229 & 265 \\
        & \# Edges  & 8,158 & 216,123 & 74,420 & 1,554,578 & 358,574 & 500,939 & 292 & 310 & 394 & 510  \\
        & Node homophily\tnote{a}  & 0.810 & 0.659 & 0.790 & 0.464 & 0.784 & 0.790 & 0.116 & 0.067 & 0.162 & 0.151 \\
        & Noise rate\tnote{b} & 0.306 & 0.309 & 0.303 & 0.575 & 0.322 & 0.356 & 0.272 & 0.246 & 0.314 & 0.283\\
        
        \midrule
        
        \multirow{7}{*}{\rotatebox{90}{\textbf{Noise types}}} 
       
        & Clean & 87.0 \scriptsize{$\pm$0.99} & 78.4 \scriptsize{$\pm$0.48} & 81.6 \scriptsize{$\pm$0.33} & 49.6 \scriptsize{$\pm$0.39} & 83.2 \scriptsize{$\pm$0.51} & 73.3 \scriptsize{$\pm$0.11} & 76.8 \scriptsize{$\pm$6.64} & 79.7 \scriptsize{$\pm$6.31} & 72.9 \scriptsize{$\pm$7.56} & 81.9 \scriptsize{$\pm$4.93} \\

        & Topology & 80.2 \scriptsize{$\pm$1.47} & 74.7 \scriptsize{$\pm$1.37} & 79.7 \scriptsize{$\pm$0.45} & 46.5 \scriptsize{$\pm$0.33} & 81.7 \scriptsize{$\pm$0.44} & 73.6 \scriptsize{$\pm$0.15} & 60.8 \scriptsize{$\pm$7.48} & 65.9 \scriptsize{$\pm$7.66} & 60.0 \scriptsize{$\pm$6.59} & 64.2 \scriptsize{$\pm$7.55} \\

        & Feature & 77.2 \scriptsize{$\pm$0.90} & 69.0 \scriptsize{$\pm$0.89} & 72.1 \scriptsize{$\pm$0.87} & 47.9 \scriptsize{$\pm$0.43} & 82.4 \scriptsize{$\pm$0.47} & 72.2 \scriptsize{$\pm$0.43} & 70.3 \scriptsize{$\pm$6.66} & 69.7 \scriptsize{$\pm$8} & 73.1 \scriptsize{$\pm$6.08} & 72.5 \scriptsize{$\pm$7.7} \\

        & Confidence & 76.7 \scriptsize{$\pm$1.12} & 74.1 \scriptsize{$\pm$0.63} & 74.8 \scriptsize{$\pm$0.00} & 46.4 \scriptsize{$\pm$0.49} & 81.5 \scriptsize{$\pm$0.49} & 70.0 \scriptsize{$\pm$0.00} & 66.6 \scriptsize{$\pm$5.89} & 67.3 \scriptsize{$\pm$8.75} & 65.1 \scriptsize{$\pm$6.6} & 70.4 \scriptsize{$\pm$6.91} \\

        & Uniform & 73.6 \scriptsize{$\pm$3.19} & 70.0 \scriptsize{$\pm$2.38} & 76.5 \scriptsize{$\pm$1.15} & 47.9 \scriptsize{$\pm$0.36} & 82.4 \scriptsize{$\pm$0.47} & 72.3 \scriptsize{$\pm$0.44} & 61.6 \scriptsize{$\pm$4.88} & 67.8 \scriptsize{$\pm$4.59} & 67.3 \scriptsize{$\pm$6.74} & 70.2 \scriptsize{$\pm$4.61} \\

        & Pairwise & 66.5 \scriptsize{$\pm$3.33} & 66.2 \scriptsize{$\pm$2.62} & 76.0 \scriptsize{$\pm$0.99} & 12.0 \scriptsize{$\pm$0.57} & 78.8 \scriptsize{$\pm$0.62} & 66.1 \scriptsize{$\pm$0.67} & 57.4 \scriptsize{$\pm$11.0} & 64.9 \scriptsize{$\pm$6.73} & 60.7 \scriptsize{$\pm$8.02} & 69.2 \scriptsize{$\pm$7.21} \\

        & LLM & 70.3 \scriptsize{$\pm$1.27} & 65.1 \scriptsize{$\pm$1.25} & 65.5 \scriptsize{$\pm$0.58} & 39.2 \scriptsize{$\pm$0.44} & 66.5 \scriptsize{$\pm$0.38} & 59.3 \scriptsize{$\pm$0.04} & 62.4 \scriptsize{$\pm$7.26} & 58.6 \scriptsize{$\pm$9.9} & 52.0 \scriptsize{$\pm$7.84} & 57.7 \scriptsize{$\pm$6.38} \\
        \bottomrule
        
    \end{tabular}
    \begin{tablenotes}       
    \item[] $\prescript{\mathrm{a}}{}{\text{It is the ratio of node’s neighbors that share the same class label.}}$ $\prescript{\mathrm{b}}{}{\text{The noise rates obtained from LLM-based noise are identically applied across all noise types.}}$
    \item[]  $\prescript{\mathrm{c}}{}{\text{Following previous research~\cite{zheng2022graph}, we categorize datasets with low node homophily as heterophily.}}$

\end{tablenotes}
\end{threeparttable}    
    }

        \vspace{-0.3cm}
\end{table*}

\begin{table}[ht]
    \centering
    \renewcommand{\arraystretch}{1.2} 
    \small 
 \caption{Example of an LLM-generated response for the History dataset (True label: World).}
    \begin{tabular}{  p{7.9cm} }
        
        \hline
            \textbf{Title}: Islam: A Concise Introduction  
        
        \textbf{Description}:   
        \textit{``Huston Smith is internationally known and revered as the premier teacher of world religions. He is the focus of a five-part PBS television series with Bill Moyers and has taught at Washington University, the Massachusetts Institute of Technology, Syracuse University, and the University of California at Berkeley. The recipient of twelve honorary degrees, Smith's fifteen books include his bestselling The World's Religions, Why Religion Matters, and his autobiography, Tales of Wonder.''}. \\

        \hline
        \textbf{LLM Response}:  Middle East\\
        This book focuses on Islam, a major world religion that originated in the Middle East, and is authored by a prominent scholar in the field of world religions, making it most relevant to the Middle East category.\\ 
        \hline
    \end{tabular}

    \label{tab:llm_response}
    \vspace{-0.3cm}
\end{table}

\subsection{LLM-based Instance-dependent Label Noise}
\label{sec:llm_noise}

While the algorithmic approach provides a systematic framework for generating instance-dependent noise, it may still not fully reflect the labeling process in real-world applications.
Therefore, we further leverage large language models (LLMs) to generate realistic label noise. 
By learning from vast amounts of human-generated text, LLMs have demonstrated high effectiveness in simulating human decision-making processes~\cite{gilardi2023chatgpt, laskar2023can, tan2024large, ahmed2024can, pavlovic2024effectiveness, kumar2024selective}.
Recent studies have shown that LLMs often exhibit error patterns similar to those of humans~\cite{wang2024resilience, havrilla2024understanding}, including systematic biases, overgeneralization, and sensitivity to ambiguous wording.

Using LLMs, we generate realistic instance-dependent noise for graph data, particularly where textual content plays a crucial role in labeling~\cite{mccallum2000automating, chiang2019cluster, wang2020microsoft, mernyei2020wiki, pei2020geom}. 
Specifically, we employ GPT-4o-mini, leveraging its strong reasoning capabilities. 
Refer to the Appendix \ref{apx:prompt} for the detailed prompt.
Table \ref{tab:llm_response} presents an example of an LLM-generated response for the History dataset, where nodes correspond to books and labels represent 12 book categories.
The LLMs misclassified \textit{Islam: A Concise Introduction} as ``Middle East'' instead of ``World'', likely due to semantic overlap between Islam and the Middle East.
This error closely resembles real-world annotation mistakes, where contextual associations influence mislabeling.

\section{Benchmark and Analysis of \\Label Noise in Graph Data}

We introduce a new benchmark for label noise in graphs by applying the proposed corruption process with various noise types to $10$~existing graph datasets.
They encompass real-world graphs with diverse characteristics, such as the number of nodes, average degrees, and homophily ratios.
They also span multiple domains, such as citation networks, e-commerce, and web pages.
Table~\ref{tab:dataset} reports the detailed statistics.
For further details of the dataset creation, please refer to Appendix \ref{apx:dataset} and \ref{apx:prompt}.
In the following subsections, we provide extensive quantitative and exploratory analyses to comprehend the behavior of GNNs on noisy graph datasets.

\begin{figure}[tbp]
    \centering
    \includegraphics[scale=0.45]{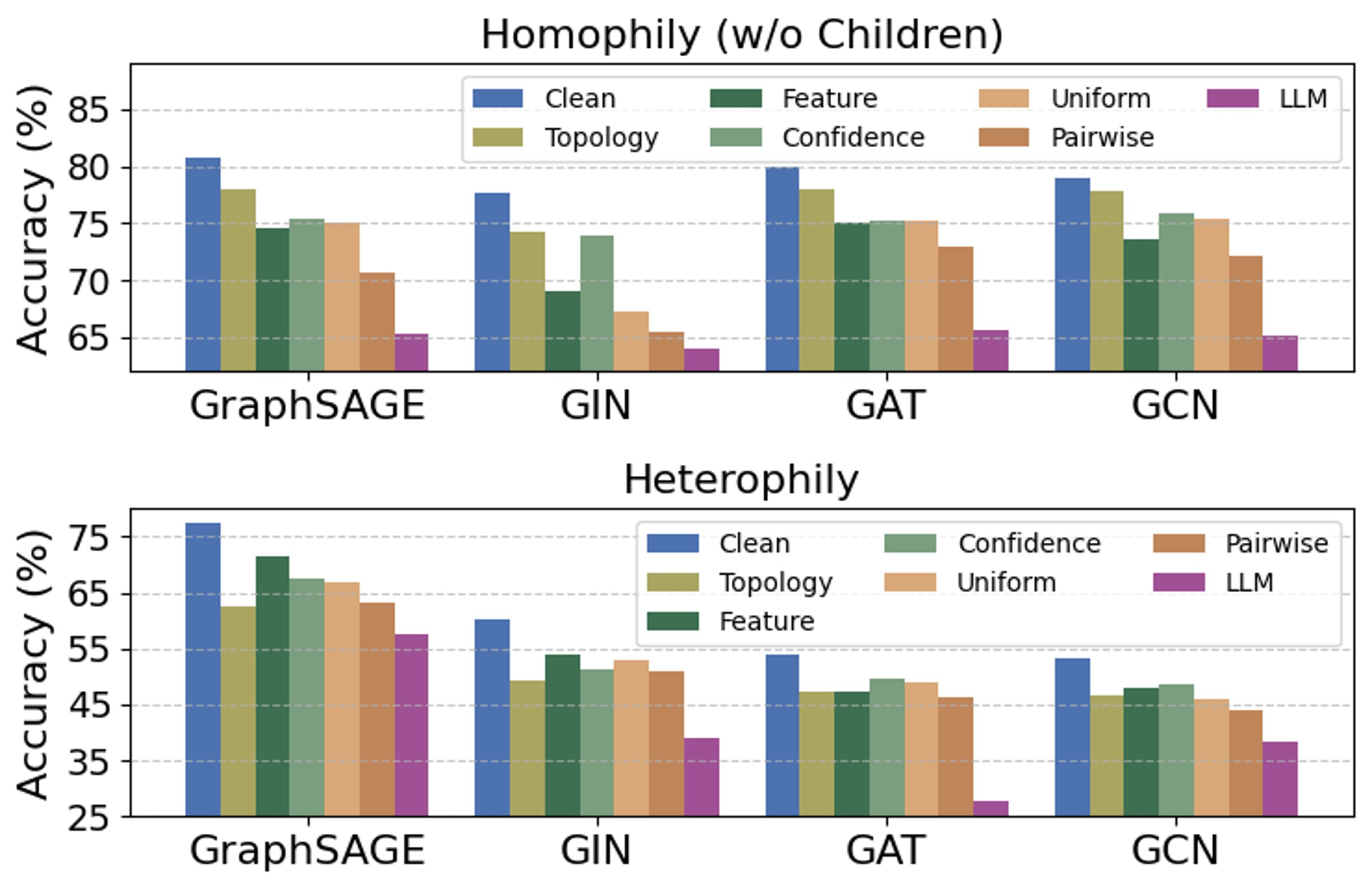}
    \caption{Average node classification accuracy for homophily and heterophily datasets over various noise types and GNNs\protect\footnotemark.}
    \label{fig:com_nois_gnn}
\end{figure}

\footnotetext{The Children dataset is excluded from homophilic accuracy calculations due to its excessive noise rate (>0.5), making it impractical for pairwise noise.}

\subsection{What is the impact of different label noise types on GNN models?}
\label{sec:label_noise_impact}
We assess the impact of various noise types on node classification accuracy in Table~\ref{tab:dataset}.
We also compare the robustness of GNN models by presenting average accuracy
across datasets in Figure~\ref{fig:com_nois_gnn}.
Overall, label noise significantly degrades node classification accuracy across various GNN models.
Our key observations are as follows:

\textbf{(1)~LLM-based label noise leads to significant performance degradation}.
This is particularly noteworthy given that the noise ratio remains the same across all noise types.
This suggests that LLM-generated noise introduces complex patterns that pose greater challenges for existing GNN models.

\textbf{(2)~Instance-dependent label noise behaves differently depending on the graph characteristics.}
Specifically, topology-based noise has a milder effect on homophilic graphs but poses more significant challenges for heterophilic ones, whereas feature-dependent noise has the opposite impact. 
In homophilic graphs, the model relies on topological consistency, while in heterophilic graphs, it prioritizes node features. 
The model remains robust by leveraging the most reliable aspect—topology for homophilic settings and features for heterophilic settings—even when noise is introduced to each respective aspect.

\textbf{(3)~Among various GNN models, GraphSAGE~\cite{kipf2016semi} achieves the best overall performance}, particularly in heterogeneous graphs where label agreement between neighboring nodes is lower.
In such cases, the standard neighborhood aggregation strategy in GNNs becomes less effective, sometimes allowing Multi-Layer Perceptron (MLP)~\cite{rumelhart1986learning} models, which rely solely on node features, to outperform GNNs (reported in Table~\ref{tab:robust_exp}).
GraphSAGE can address this issue by assigning separate parameters to the central and neighboring nodes before aggregation, allowing the model to selectively extract relevant information from the central node and reduce error propagation.
\vspace{\baselineskip}  

\subsection{How does instance-dependent noise differ from class-dependent noise?}

\begin{figure}[t]
	\centering
	\begin{subfigure}{0.49\columnwidth}
	    \centering
	    \includegraphics[width=\textwidth]{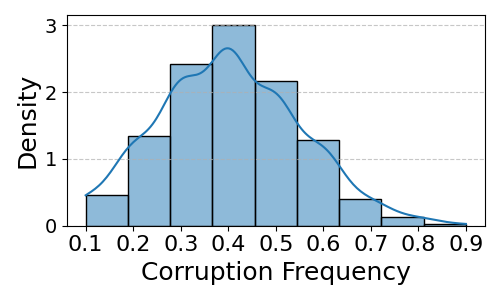}
	    \caption{Class-dependent (Uniform)}
	\end{subfigure}
	\hfill
    \begin{subfigure}{0.49\columnwidth}
	    \centering
	    \includegraphics[width=\textwidth]{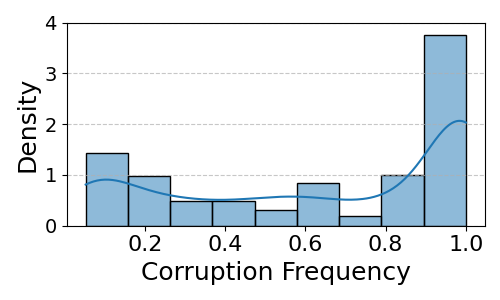}
	    \caption{Instance-dependent (LLM)}
	\end{subfigure}

\caption{The distribution of instances' corruption frequency in the Cora-ML dataset across 10 generated noisy label sets.}
\label{fig:corrup_fre}
    \vspace{-0.1cm}
\end{figure}

\begin{figure}[tbp]
    \centering
    \includegraphics[scale=0.35]{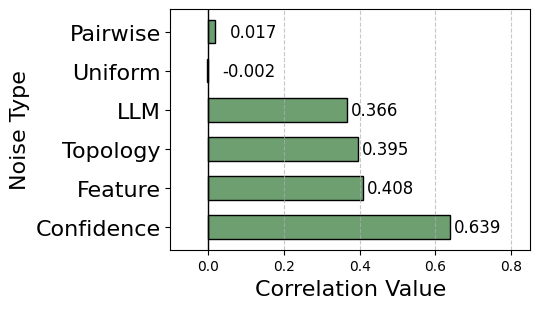}
    \caption{Correlation between the model prediction entropy and the corruption frequency on the Cora-ML dataset.}
    \label{fig:corr_hardnss}
    \vspace{-0.3cm}
\end{figure}

All prior works on label noise in graphs~\cite{wang2024noisygl,cheng2024resurrecting, zhu2024robust, du2021noise, li2024contrastive} have utilized class-dependent noise.
We explore how the instance-dependent noise introduced in this work differs from previous class-dependent noise in terms of corruption patterns and model learning.
To further analyze these differences, we conduct 10 independent corruption processes for each noise type and compute the corruption frequency of each instance, measuring how often it is mislabeled across multiple noise realizations. 
Figure \ref{fig:corrup_fre} shows the distributions of instances' corruption frequency from class-dependent (Uniform) and instance-dependent (LLM-based) noise.
Class-dependent noise follows a normal distribution with no selection patterns, applying uniform mislabeling across all instances in a class.
In contrast, \textbf{instance-dependent noise exhibits a skewed distribution, repeatedly corrupting the same instances.}

Next, we explore whether frequently corrupted instances tend to make the model more confused.
We measure model prediction uncertainty using the entropy of the output distribution, defined as: $H = - \sum_{c=1}^{C} f_c(G, Y) \log f_c(G, Y)$, where $f_c(G, Y)$ is the prediction probability for class $c$.
A higher value indicates greater ambiguity, often corresponding to more challenging instances~\cite{wei2021learning}.
Figure~\ref{fig:corr_hardnss} presents the correlation between the prediction entropy and corruption frequency.
Overall, instance-dependent noise exhibits a strong positive correlation with entropy, indicating that \textbf{the model tends to be more uncertain about frequently corrupted instances.}

Interestingly, the corruption of hard-to-classify samples does not necessarily impose a greater burden on the model. 
The confidence-based noise shows the highest correlation value, which is expected given that it directly utilizes predicted probabilities.
However, despite its high correlation, confidence-based noise is not necessarily the most challenging for model training, as shown in Figure \ref{fig:com_nois_gnn}.
This observation aligns with recent findings in noise-robust learning~\cite{oyen2022robustness}; samples near the decision boundary can be ambiguous for both humans and models, making them more prone to mislabeling~\cite{wei2021learning}. 
However, since these samples naturally exist in an uncertain region, their corruption may have a limited impact on overall performance compared to noise with a more structured bias.

\subsection{Why is LLM-based label noise challenging?}
\label{sec:LLM_challenge}

\begin{figure}[t]
    \centering
    \begin{subfigure}{0.3\linewidth}
        \centering
        \includegraphics[width=\textwidth]{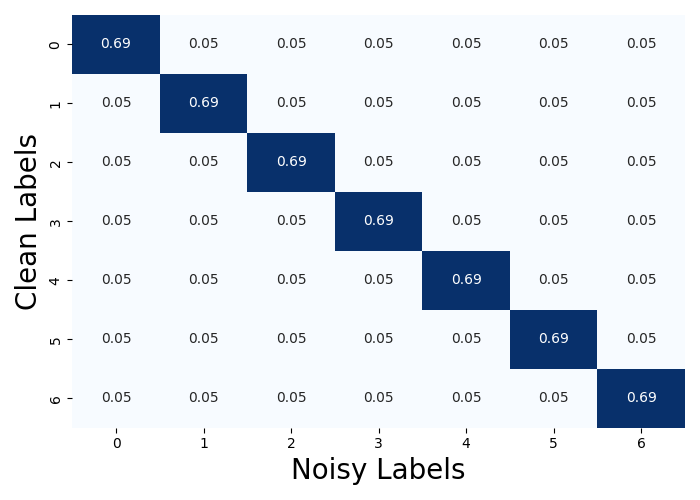}
        \caption{Uniform}
    \end{subfigure}
    \hfill
    \begin{subfigure}{0.3\linewidth}
        \centering
        \includegraphics[width=\textwidth]{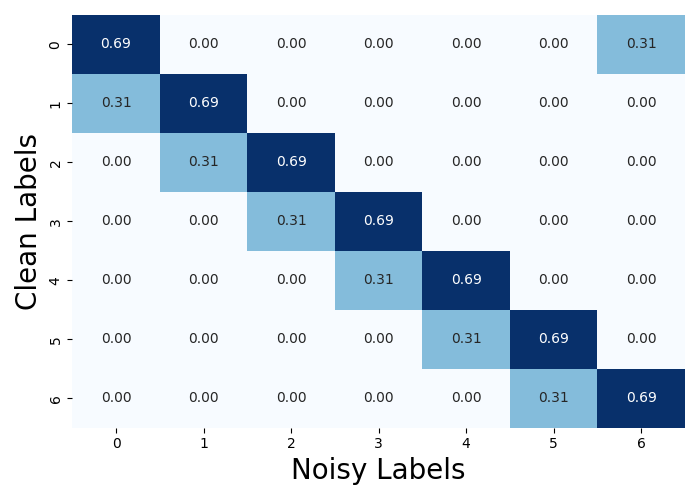}
        \caption{Pairwise}
    \end{subfigure}
    \hfill
    \begin{subfigure}{0.3\linewidth}
        \centering
        \includegraphics[width=\textwidth]{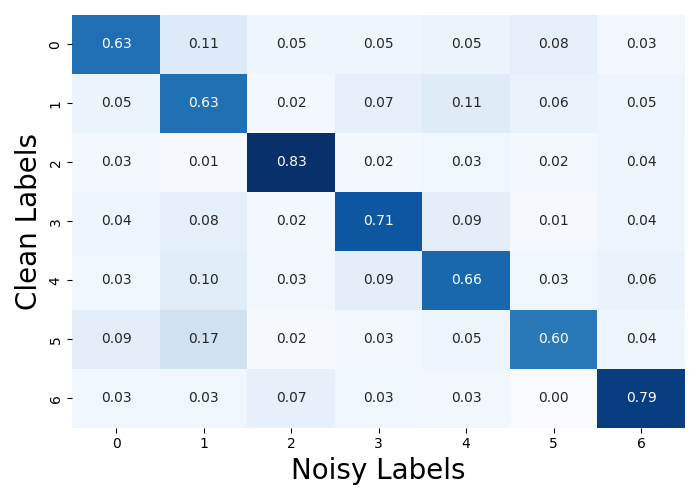}
        \caption{Topology}
    \end{subfigure}

    \vspace{0.3cm} 

    \begin{subfigure}{0.3\linewidth}
        \centering
        \includegraphics[width=\textwidth]{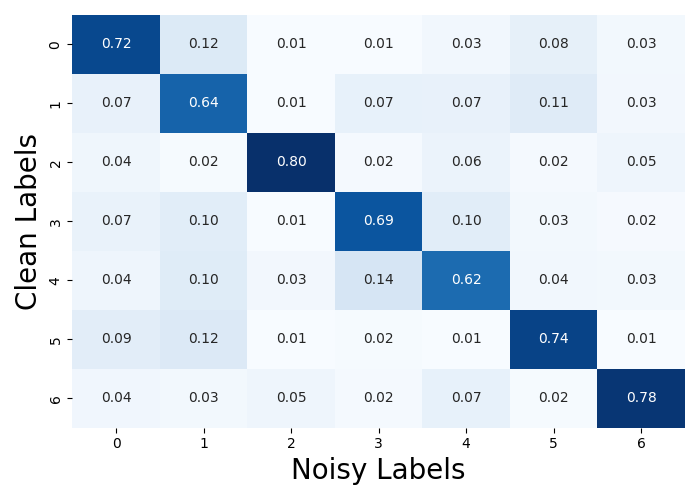}
        \caption{Feature}
    \end{subfigure}
    \hfill
    \begin{subfigure}{0.3\linewidth}
        \centering
        \includegraphics[width=\textwidth]{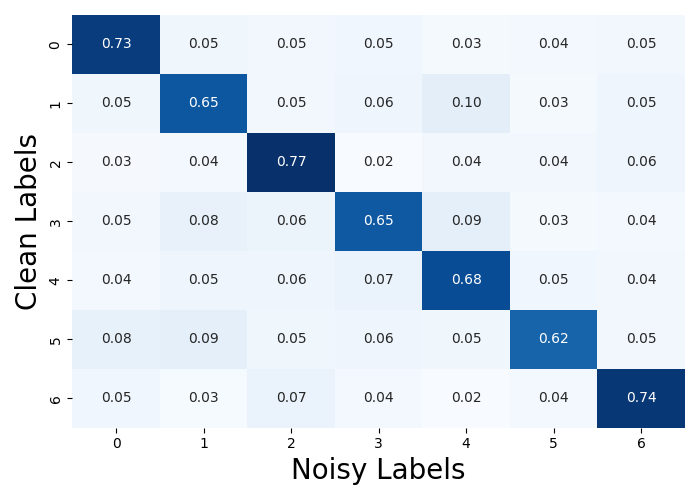}
        \caption{Confidence}
    \end{subfigure}
    \hfill
    \begin{subfigure}{0.3\linewidth}
        \centering
        \includegraphics[width=\textwidth]{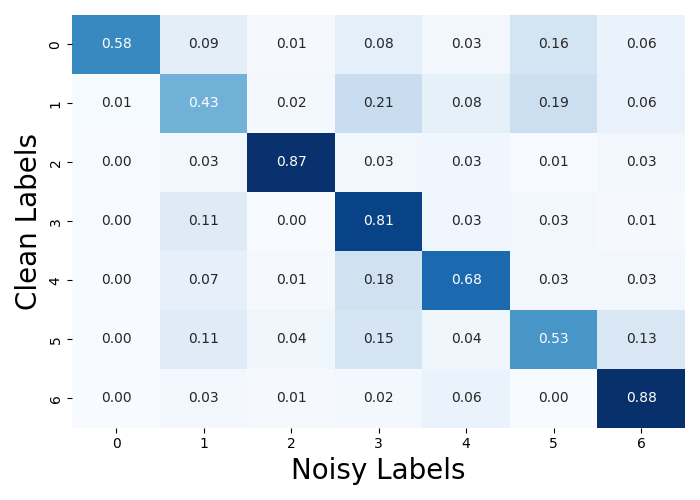}
        \caption{LLM}
    \end{subfigure}

    \caption{Transition matrices of different noise types.}
    \label{fig:tm}
    \vspace{-0.3cm} 
\end{figure}

\begin{table*}[ht]
    \centering
    \caption{Noisy label detection score (ROC-AUC, \%) on datasets with LLM-based label noise using Loss trajectory-based GMM model and supervised detector. Results are mean $\pm$ standard deviation over 10 runs, with the best in each method bolded.}
    \label{tab:detection}
        
        \resizebox{0.99\textwidth}{!}{%
        \begin{tabular}{lcccccccccccc|c}
        \toprule
         \multicolumn{2}{c}{Method} & Models & Cora-ML & WikiCS & Product-s & Children & History & Photo & Cornell & Texas & Washington & Wisconsin & Avg. \\
        \midrule  
        \multirow{8}{*}{\rotatebox{90}{\textbf{\shortstack{Loss trajectory-based}}}}

        & \multirow{4}{*}{\rotatebox{90}{\textbf{Average}}} 
        & MLP~\cite{rumelhart1986learning} & 62.5 \scriptsize{$\pm$1.21} & 55.6 \scriptsize{$\pm$0.14} & 68.2 \scriptsize{$\pm$0.09} & 55.5 \scriptsize{$\pm$0.10} & 71.6 \scriptsize{$\pm$0.40} & \textbf{56.3} \scriptsize{$\pm$0.31} & 55.6 \scriptsize{$\pm$3.86} & 51.8 \scriptsize{$\pm$2.08} & \textbf{49.0} \scriptsize{$\pm$5.1} & \textbf{59.4} \scriptsize{$\pm$1.82}& 58.6  \\

        & &GCN~\cite{hamilton2017inductive}
        & \textbf{78.0} \scriptsize{$\pm$0.12} & 63.8 \scriptsize{$\pm$0.72} & \textbf{74.1} \scriptsize{$\pm$0.10} & 51.3 \scriptsize{$\pm$0.52} & 75.5 \scriptsize{$\pm$0.25} & 54.8 \scriptsize{$\pm$1.27} & 51.4 \scriptsize{$\pm$1} & 37.9 \scriptsize{$\pm$5.18} & 35.7 \scriptsize{$\pm$6.75} & 45.5 \scriptsize{$\pm$0.31}& 56.8  \\

        & &GAT~\cite{velivckovic2017graph} & 72.0 \scriptsize{$\pm$0.26} & \textbf{65.9} \scriptsize{$\pm$0.36} & 68.6 \scriptsize{$\pm$0.67} & 54.6 \scriptsize{$\pm$2.94} & \textbf{76.9} \scriptsize{$\pm$0.45} & 54.2 \scriptsize{$\pm$0.22} & 42.2 \scriptsize{$\pm$9} & 42.6 \scriptsize{$\pm$4.47} & 39.1 \scriptsize{$\pm$1.64} & 49.0 \scriptsize{$\pm$1.97}& 56.5  \\

        & &GraphSAGE~\cite{kipf2016semi} &  69.1 \scriptsize{$\pm$0.67} & 61.9 \scriptsize{$\pm$0.94} & 64.7 \scriptsize{$\pm$0.74} & \textbf{57.3} \scriptsize{$\pm$0.64} & 75.8 \scriptsize{$\pm$1.89} & 54.1 \scriptsize{$\pm$2.42} & \textbf{55.8} \scriptsize{$\pm$0.97} & \textbf{58.2} \scriptsize{$\pm$2.47} & 45.5 \scriptsize{$\pm$0.98} & 55.3 \scriptsize{$\pm$0.91}& \textbf{59.8}  \\

        \arrayrulecolor{gray} 
        \cmidrule(lr){2-14} 
        \arrayrulecolor{black}

        & \multirow{4}{*}{\rotatebox{90}{\textbf{Maximum}}}      
        & MLP~\cite{rumelhart1986learning} & 65.4 \scriptsize{$\pm$0.52} & 59.4 \scriptsize{$\pm$0.11} & 66.9 \scriptsize{$\pm$0.73} & 55.6 \scriptsize{$\pm$0.33} & 71.2 \scriptsize{$\pm$0.34} & 58.2 \scriptsize{$\pm$0.16} & 72.4 \scriptsize{$\pm$0.67} & 69.5 \scriptsize{$\pm$1.91} & 64.9 \scriptsize{$\pm$4.75} & 72.4 \scriptsize{$\pm$1.98}& 65.6  \\

        & & GCN~\cite{hamilton2017inductive} & \textbf{78.6} \scriptsize{$\pm$0.12} & 73.1 \scriptsize{$\pm$1.00} & 70.9 \scriptsize{$\pm$0.15} & 58.6 \scriptsize{$\pm$1.54} & 76.4 \scriptsize{$\pm$1.11} & 59.3 \scriptsize{$\pm$1.06} & 68.7 \scriptsize{$\pm$9.78} & \textbf{82.3} \scriptsize{$\pm$0.91} & 67.4 \scriptsize{$\pm$7.25} & 50.5 \scriptsize{$\pm$0.78}& 68.6  \\

        & & GAT~\cite{velivckovic2017graph} & 57.3 \scriptsize{$\pm$0.41} & 69.8 \scriptsize{$\pm$0.18} & 71.2 \scriptsize{$\pm$0.58} & 55.8 \scriptsize{$\pm$0.94} & 76.1 \scriptsize{$\pm$1.57} & 55.9 \scriptsize{$\pm$2.08} & 55.2 \scriptsize{$\pm$1.58} & 56.9 \scriptsize{$\pm$2.4} & 56.2 \scriptsize{$\pm$3.35} & 51.7 \scriptsize{$\pm$0.85}& 60.6  \\

        & & GraphSAGE~\cite{kipf2016semi} & 74.0 \scriptsize{$\pm$1.3} & \textbf{74.1} \scriptsize{$\pm$1.74} & \textbf{75.0} \scriptsize{$\pm$1.11} & \textbf{61.3} \scriptsize{$\pm$1.54} & \textbf{79.5} \scriptsize{$\pm$1.42} & \textbf{63.1} \scriptsize{$\pm$0.93} & \textbf{79.8} \scriptsize{$\pm$2.13} & 72.0 \scriptsize{$\pm$4.69} & \textbf{82.6} \scriptsize{$\pm$1.07} & \textbf{83.6} \scriptsize{$\pm$1.96}& \textbf{74.5}  \\

        \midrule

        \multirow{4}{*}{\rotatebox{90}{\textbf{Detector}}} 
        
        && MLP~\cite{rumelhart1986learning} & 74.5 \scriptsize{$\pm$2.99} & 76.6 \scriptsize{$\pm$2.11} & 81.5 \scriptsize{$\pm$0.63} & \textbf{68.6} \scriptsize{$\pm$0.64} & \textbf{85.9} \scriptsize{$\pm$0.74} & 73.6 \scriptsize{$\pm$0.66} & \textbf{70.6} \scriptsize{$\pm$15.2} & \textbf{75.6} \scriptsize{$\pm$11.4} & \textbf{72.5} \scriptsize{$\pm$11.8} & \textbf{85.9} \scriptsize{$\pm$7.25}& \textbf{76.5}  \\

        &&GCN~\cite{hamilton2017inductive}
         & 76.2 \scriptsize{$\pm$3.28} & 79.0 \scriptsize{$\pm$1.7} & 79.5 \scriptsize{$\pm$0.30} & 64.6 \scriptsize{$\pm$0.51} & 85.2 \scriptsize{$\pm$0.56} & 72.8 \scriptsize{$\pm$0.63} & 55.3 \scriptsize{$\pm$6.52} & 51.5 \scriptsize{$\pm$16.4} & 60.2 \scriptsize{$\pm$12.2} & 65.1 \scriptsize{$\pm$11.3}& 68.9  \\

        &&GAT~\cite{velivckovic2017graph} & 74.3 \scriptsize{$\pm$2.85} & \textbf{79.9} \scriptsize{$\pm$1.55} & \textbf{86.3} \scriptsize{$\pm$0.45} & 60.2 \scriptsize{$\pm$1.9} & 82.9 \scriptsize{$\pm$1.79} & 74.3 \scriptsize{$\pm$0.69} & 51.0 \scriptsize{$\pm$21.3} & 50.7 \scriptsize{$\pm$13.5} & 53.1 \scriptsize{$\pm$13.9} & 46.6 \scriptsize{$\pm$14.8}& 65.9  \\

        &&GraphSAGE~\cite{kipf2016semi} &  \textbf{77.4} \scriptsize{$\pm$1.99} & 77.2 \scriptsize{$\pm$1.89} & 84.0 \scriptsize{$\pm$1.36} & 66.2 \scriptsize{$\pm$0.39} & 84.1 \scriptsize{$\pm$0.71} & \textbf{74.5} \scriptsize{$\pm$0.57} & 70.0 \scriptsize{$\pm$13.1} & 64.8 \scriptsize{$\pm$12.8} & 72.1 \scriptsize{$\pm$14.1} & 85.2 \scriptsize{$\pm$6.1}& 75.6  \\

        \bottomrule
    \end{tabular}
    }
    \vspace{-0.2cm}
\end{table*}

Table~\ref{tab:dataset} and Figure~\ref{fig:com_nois_gnn} consistently show that LLM-based noise is particularly challenging for GNNs to handle.
To gain deeper insights into this phenomenon, we conduct further analysis.
In Figure \ref{fig:tm}, we present transition matrices from different types of label noise on the Cora-ML dataset.
We first observe that instance-dependent label noise exhibits more diverse transition patterns compared to class-dependent label noise.
Next, we compare LLM-based transition patterns with those of other noise types.
Compared to uniform noise, LLM-based noise tends to mislead instances in a more structured manner, guiding them toward specific classes influenced by instance-level uncertainties.
At first glance, topology- and feature-dependent noise may appear similar to LLM-based noise. 
However, upon closer examination, we find that LLM-based noise produces more uneven transition patterns among classes.

\begin{figure}[t]
    \centering
    \includegraphics[scale=0.35]{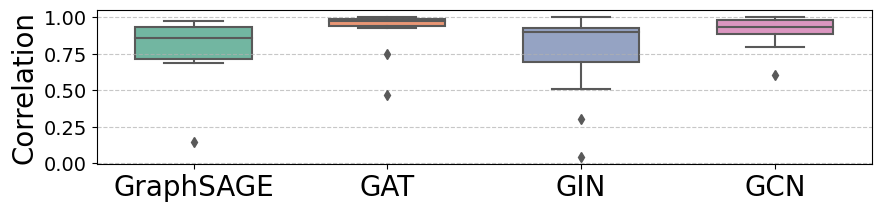}
    \caption{Correlation between the entropy of off-diagonal values in transition matrices and model performance. }
    \label{fig:tm_corr}
        \vspace{-0.5cm}
\end{figure}

To assess the impact of these patterns on model performance, we compute the correlation between the entropy of off-diagonal values in the transition matrix and model performance across topology-, feature-, and LLM-based noise types for our benchmark datasets.
Here, higher entropy indicates more uniform transitions among classes.
Figure~\ref{fig:tm_corr} shows a strong positive correlation between entropy values and model performance, implying that lower performance is associated with a more skewed distribution of off-diagonal probabilities.
In other words, \textbf{a model tends to perform worse when transitions are concentrated in a few specific classes} rather than being evenly distributed across multiple classes.
One noteworthy finding is that \textbf{this effect becomes more pronounced when instance-level uncertainty is reflected in the corruption process.}
While pairwise noise, which flips labels based on predefined class pairs, exhibits the lowest entropy values, its corruption process is class-dependent and ignores instance-level uncertainties.
This appears to make learning for GNNs relatively easier compared to the instance-dependent noise introduced by LLMs.

Lastly, despite directly incorporating prediction confidence into the corruption process, confidence-based noise exhibits more uniform off-diagonal values than other instance-dependent noise.
We observe that GNNs tend to be overly confident in a single class, which reduces the discriminative power of corruption probabilities across instances and results in a less structured noise distribution.

In sum, we conclude that LLM-based noise is particularly challenging to handle, as it generates mislabeling toward a few contextually plausible alternative classes by incorporating instance-dependent uncertainties. 
This can make GNN models more susceptible to learning incorrect patterns.

\section{Benchmark and Analysis of \\Label Noise Handling in Graph Data}
\label{sec:Label Noise handle for Graph Data}

To address noisy datasets, two well-established approaches can be considered: 
(1) detecting corrupted instances to exclude the nodes from training or relabel them, 
or (2) utilizing special learning techniques or model architectures designed to maintain strong generalization while being robust to corrupted instances.
We provide a detailed discussion of both approaches. 
In this section, we utilize LLM-based label noise, which GNNs have consistently struggled with the most, as shown in the previous section.

\subsection{Noisy Label Detection in Graph Data}

Effective detection of corrupted instances relies on identifying distinctive patterns that differentiate them from clean ones.  
We first analyze the effectiveness of the loss trajectory throughout the training process (Sec. \ref{sec:detector_loss}) and evaluate the discriminative power of GNNs by devising a supervised noisy label detector (Sec. \ref{sec:detector_sup}).
We then discuss the applicability of two distinct types of information: feature similarity and topology-based label consistency (Sec. \ref{sec:detector_feattopo}).

\subsubsection{\textbf{Loss trajectory-based noisy label detection}}
\label{sec:detector_loss}

\begin{figure}[t]
	\centering
	\begin{subfigure}{0.32\columnwidth}
	    \centering
	    \includegraphics[width=\textwidth]{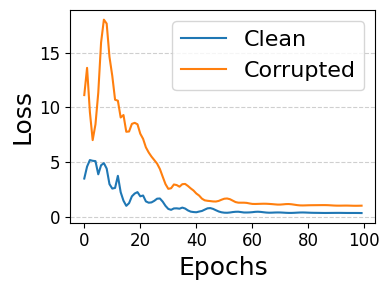}
	    \caption{Loss trajectory }
        \label{fig:loss_traj_a}
	\end{subfigure}
	\hfill
    \begin{subfigure}{0.32\columnwidth}
	    \centering
	    \includegraphics[width=\textwidth]{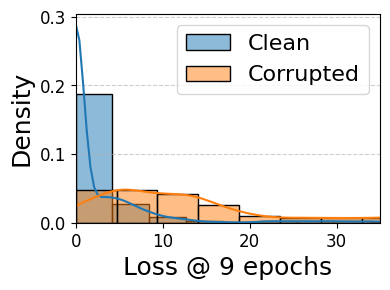}
	    \caption{Loss distribution}
        \label{fig:loss_traj_b}
	\end{subfigure}
    \begin{subfigure}{0.32\columnwidth}
	    \centering
	    \includegraphics[width=\textwidth]{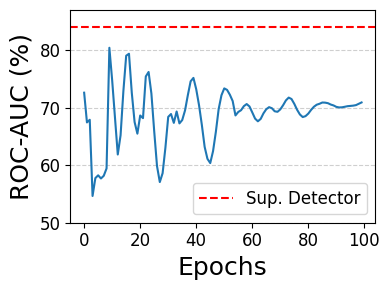}
	    \caption{Detection score}
        \label{fig:loss_traj_c}
	\end{subfigure}

\caption{Training losses and loss-based detection scores on the History dataset using GraphSAGE.   }
\label{fig:loss_traj}
    \vspace{-0.3cm}

\end{figure}

A widely adopted technique, known as the small-loss trick~\cite{han2018co, huang2019o2u, li2020dividemix}, assumes that instances with higher loss are more likely to be corrupted, particularly in the early stages of training. 
Figure \ref{fig:loss_traj_a} presents the average loss trajectories of clean and corrupted instances over training epochs, where clean instances quickly converge to lower values, making them distinguishable from corrupted instances.
Similarly, Figure \ref{fig:loss_traj_b} presents the loss distribution at epoch 9, showing that clean instances exhibit significantly lower loss values.

By leveraging these differences, we can distinguish clean from corrupted instances using a Gaussian Mixture Model (GMM)~\cite{zoran2011learning}.
Figure \ref{fig:loss_traj_c} presents how detection performance evolves over training epochs.
While this loss-based detection achieves considerable effectiveness, it also exhibits significant fluctuations in detection scores over training epochs.
This highlights the need for strategies to extract stable and meaningful signals by leveraging techniques like early stopping~\cite{bai2021understanding} and loss dynamics modeling~\cite{kim2024learning} to improve detection performance.

In Table \ref{tab:detection}, we report loss trajectory-based detection performance across various backbone models under two setups:
(1) ``Average'' refers to a single evaluation based on the average loss over all epochs, making it the most straightforward and naive approach.
(2) ``Maximum'' refers to assessing performance at each epoch and reporting the highest observed value.
These results serve as a baseline for detection performance, providing a reference point for evaluating more advanced detection methods in future work.

\subsubsection{\textbf{Supervised detector-based noisy label detection}}   
\label{sec:detector_sup}
To assess how effectively GNNs capture discriminative patterns in corrupted instances, we design a controlled experiment setting, assuming the availability of clean labels for the training dataset.
We formulate the problem of detecting noisy labels as a binary classification task.
Given \( D = \{G, \mathrm{Y}, \tilde{\mathrm{Y}}\} \), we construct the dataset for the detector as: \( D^{det} = \{G, \mathrm{Y}^{det}\} \), where $\mathrm{Y}^{det} = \mathbbm{1} [ \mathrm{\mathrm{Y}} \neq \tilde{\mathrm{Y}}]$, labeling the corrupted instances as 1 and the clean ones as 0.
We split the dataset into training, validation, and test sets in an 8:1:1 ratio and train the detectors until the validation~performance~peaks.

\begin{table*}[h]
    \centering
    \caption{Node classification accuracy (\%) on datasets with LLM-based label noise across various models. Results are mean $\pm$ standard deviation over 10 runs. For each dataset, the best value is bolded and underlined, while the top three are bolded. LLN, G-LLN and OOM indicate Learning with Label Noise, GNN integrated LLN, and out of memory, respectively. }
    \label{tab:robust_exp}
        
        \resizebox{0.99\textwidth}{!}{%
        \begin{tabular}{lccccccc|cccc|ccc}
        \toprule

        & &  \multicolumn{6}{c|}{Homophily datasets} & \multicolumn{4}{c|}{Heterophily datasets} & \multicolumn{3}{c}{Average} \\
        & Dataset  & Cora-ML & WikiCS & Product-s & Children & History & Photo & Cornell & Texas & Washington & Wisconsin & Homo. & Hetero. & All \\
        & Noise rate  & 0.306 & 0.309 & 0.303 & 0.575 & 0.322 & 0.356 & 0.272 & 0.246 & 0.314 & 0.283 & 0.362 &	0.279 & 0.329 \\
    
        \midrule        

         \multirow{6}{*}{\rotatebox{90}{\textbf{Base models}}}  & MLP~\cite{rumelhart1986learning} & 65.8 \scriptsize{$\pm$1.25} & 60.7 \scriptsize{$\pm$0.07} & 54.2 \scriptsize{$\pm$0.27} & 37.8 \scriptsize{$\pm$0.98} & 65.7 \scriptsize{$\pm$0.57} & 55.0 \scriptsize{$\pm$0.58} & \textbf{56.1} \scriptsize{$\pm$4.71} & \textbf{58.6} \scriptsize{$\pm$7.93} & \textbf{49.1 }\scriptsize{$\pm$6.48} & \textbf{60.4} \scriptsize{$\pm$7.5}&56.5&\textbf{56.0} &\textbf{56.3}\\


        &GCN~\cite{hamilton2017inductive} &  71.5 \scriptsize{$\pm$1.2} & \textbf{67.1} \scriptsize{$\pm$1.1} & 61.3 \scriptsize{$\pm$0.52} & 37.1 \scriptsize{$\pm$0.83} & \textbf{\underline{66.6}} \scriptsize{$\pm$0.41} & 59.4 \scriptsize{$\pm$0.88} & 37.1 \scriptsize{$\pm$4.77} & 51.9 \scriptsize{$\pm$4.49} & 26.7 \scriptsize{$\pm$7.95} & 37.0 \scriptsize{$\pm$6.97}& 60.5& 38.2&51.6\\
        
        &GAT~\cite{velivckovic2017graph} & 70.3 \scriptsize{$\pm$1.36} & 67.0 \scriptsize{$\pm$0.73} & 66.6 \scriptsize{$\pm$0.35} & 31.5 \scriptsize{$\pm$0.89} & 64.7 \scriptsize{$\pm$1.21} & 59.4 \scriptsize{$\pm$1.36} & 24.2 \scriptsize{$\pm$9.25} & 41.4 \scriptsize{$\pm$11.7} & 25.3 \scriptsize{$\pm$7.52} & 19.4 \scriptsize{$\pm$5.34}& 59.9& 27.6&47.0\\
        
        &GraphSage~\cite{kipf2016semi}&  70.3 \scriptsize{$\pm$1.27} & 65.1 \scriptsize{$\pm$1.25} & 65.5 \scriptsize{$\pm$0.58} & \textbf{39.2} \scriptsize{$\pm$0.44} & \textbf{66.5} \scriptsize{$\pm$0.38} & 59.3 \scriptsize{$\pm$0.04} & \textbf{\underline{62.4}} \scriptsize{$\pm$7.26} & \textbf{58.6} \scriptsize{$\pm$9.9} & \textbf{52.0} \scriptsize{$\pm$7.84} & 57.7 \scriptsize{$\pm$6.38}& 61.0 &\textbf{57.7} &\textbf{\underline{59.7}} \\
        
        &GIN~\cite{xu2018powerful} & 69.1 \scriptsize{$\pm$3.05} & 64.5 \scriptsize{$\pm$0.24} & 59.4 \scriptsize{$\pm$0.77} & 36.6 \scriptsize{$\pm$0.49} & 66.0 \scriptsize{$\pm$0.40} & 60.8 \scriptsize{$\pm$1.14} & 44.2 \scriptsize{$\pm$9.02} & 49.5 \scriptsize{$\pm$7.05} & 33.6 \scriptsize{$\pm$8.58} & 28.9 \scriptsize{$\pm$4.55}& 59.4& 39.0 &51.2\\
        
        &LCAT~\cite{javaloy2023learnable} &  67.9 \scriptsize{$\pm$1.31} & 65.3 \scriptsize{$\pm$0.72} & 58.1 \scriptsize{$\pm$0.26} & 36.9 \scriptsize{$\pm$1.02} & 65.8 \scriptsize{$\pm$0.43} & 58.1 \scriptsize{$\pm$0.30} & 46.1 \scriptsize{$\pm$10.3} & 44.6 \scriptsize{$\pm$11.2} & \textbf{49.1} \scriptsize{$\pm$6.85} & 55.3 \scriptsize{$\pm$6.32}& 58.7& 48.8 & 54.7  \\

        \midrule
        
        \multirow{7}{*}{\rotatebox{90}{\textbf{LLN}}} 
       
        &S-model~\cite{goldberger2017training} & 73.7 \scriptsize{$\pm$1.67} & 66.5 \scriptsize{$\pm$0.69} & \textbf{66.9} \scriptsize{$\pm$0.34} & 38.5 \scriptsize{$\pm$0.24} & 66.2 \scriptsize{$\pm$0.43} & \textbf{64.3} \scriptsize{$\pm$0.46} & 40.3 \scriptsize{$\pm$6.87} & 40.8 \scriptsize{$\pm$9.4} & 32.7 \scriptsize{$\pm$8.08} & 35.8 \scriptsize{$\pm$7.01}& \textbf{62.7}& 37.4 & 52.6  \\
        
        &Forward~\cite{patrini2017making}  &  69.4 \scriptsize{$\pm$6.72} & 66.0 \scriptsize{$\pm$2.94} & 30.3 \scriptsize{$\pm$3.39} & 28.3 \scriptsize{$\pm$3.99} & 65.3 \scriptsize{$\pm$3.19} & 38.2 \scriptsize{$\pm$11.2} & 29.7 \scriptsize{$\pm$10.7} & 42.4 \scriptsize{$\pm$10.9} & 30.2 \scriptsize{$\pm$7.9} & 40.9 \scriptsize{$\pm$5.6}& 49.6& 35.8 & 44.1  \\
        
        &Backward~\cite{patrini2017making} & \textbf{74.1} \scriptsize{$\pm$1.5} & 59.8 \scriptsize{$\pm$2.59} & 5.8 \scriptsize{$\pm$2.28} & 23.8 \scriptsize{$\pm$5.53} & 58.5 \scriptsize{$\pm$16.0} & 47.9 \scriptsize{$\pm$13.9} & 41.1 \scriptsize{$\pm$6.98} & 25.7 \scriptsize{$\pm$14.3} & 20.0 \scriptsize{$\pm$11.6} & 31.3 \scriptsize{$\pm$8.32}& 45.0& 29.5 & 38.8  \\

        &Coteaching~\cite{han2018co} & 68.8 \scriptsize{$\pm$1.71} & 61.9 \scriptsize{$\pm$1.01} & \textbf{\underline{67.9}} \scriptsize{$\pm$0.56} & 33.2 \scriptsize{$\pm$7.95} & 60.1 \scriptsize{$\pm$4.67} & 30.3 \scriptsize{$\pm$17.3} & 36.1 \scriptsize{$\pm$13.2} & 37.0 \scriptsize{$\pm$12.2} & 33.6 \scriptsize{$\pm$9.66} & 38.1 \scriptsize{$\pm$9.94}& 53.7& 36.2 & 46.7  \\

        &SCE~\cite{wang2019symmetric} &  73.4 \scriptsize{$\pm$2.12} & \textbf{67.1} \scriptsize{$\pm$1.13} & 65.6 \scriptsize{$\pm$0.37} & 38.8 \scriptsize{$\pm$0.72} & 65.9 \scriptsize{$\pm$0.39} & 62.1 \scriptsize{$\pm$1.33} & 29.5 \scriptsize{$\pm$8.31} & 44.6 \scriptsize{$\pm$11.6} & 33.1 \scriptsize{$\pm$8.63} & 37.5 \scriptsize{$\pm$7.8}& \textbf{62.2}& 36.2 & 51.8  \\

        &JoCoR~\cite{wei2020combating} &\textbf{\underline{75.5}} \scriptsize{$\pm$1.85} & 57.5 \scriptsize{$\pm$1.21} & 50.7 \scriptsize{$\pm$1.08} & 36.2 \scriptsize{$\pm$1.46} & 62.3 \scriptsize{$\pm$1.13} & 46.9 \scriptsize{$\pm$1.22} & 32.4 \scriptsize{$\pm$7.72} & 36.2 \scriptsize{$\pm$11.3} & 25.1 \scriptsize{$\pm$8.67} & 34.5 \scriptsize{$\pm$6.54}& 54.9& 32.1 & 45.7  \\
        
        &APL~\cite{ma2020normalized} &72.4 \scriptsize{$\pm$1.46} & \textbf{67.1} \scriptsize{$\pm$0.54} & \textbf{67.0} \scriptsize{$\pm$0.28} & 35.6 \scriptsize{$\pm$1.52} & 62.4 \scriptsize{$\pm$0.44} & 43.2 \scriptsize{$\pm$1.21} & 37.1 \scriptsize{$\pm$6.28} & 41.1 \scriptsize{$\pm$9.35} & 27.8 \scriptsize{$\pm$9.18} & 39.4 \scriptsize{$\pm$6.34}& 58.0& 36.3 & 49.3  \\

       \midrule

        \multirow{11}{*}{\rotatebox{90}{\textbf{G-LLN}}} 
        &D-GNN~\cite{hoang2019learning} &72.9 \scriptsize{$\pm$2.26} & 64.1 \scriptsize{$\pm$0.65} & 46.4 \scriptsize{$\pm$1.86} & 35.1 \scriptsize{$\pm$1.54} & 58.4 \scriptsize{$\pm$19.5} & 60.1 \scriptsize{$\pm$1.18} & 40.8 \scriptsize{$\pm$13.5} & 41.9 \scriptsize{$\pm$16.7} & 29.1 \scriptsize{$\pm$11.0} & 12.5 \scriptsize{$\pm$14.1}& 56.2& 31.1 & 46.1  \\
        
        &CP~\cite{zhang2020adversarial} & 74.0 \scriptsize{$\pm$1.81} & 66.7 \scriptsize{$\pm$0.90} & 64.8 \scriptsize{$\pm$0.26} & 38.3 \scriptsize{$\pm$0.99} & \textbf{66.4} \scriptsize{$\pm$0.50} & \textbf{62.3} \scriptsize{$\pm$0.55} & 40.3 \scriptsize{$\pm$5.53} & 38.4 \scriptsize{$\pm$8.7} & 31.3 \scriptsize{$\pm$9.96} & 36.0 \scriptsize{$\pm$4.5}& 62.1& 36.5 & 51.8  \\
        
        &NRGNN~\cite{dai2021nrgnn} & 68.8 \scriptsize{$\pm$1.79} & 61.5 \scriptsize{$\pm$0.81} & 61.2 \scriptsize{$\pm$0.88} & \textbf{\underline{40.6}} \scriptsize{$\pm$0.46} & 64.6 \scriptsize{$\pm$0.62} & 55.2 \scriptsize{$\pm$0.55} & 50.5 \scriptsize{$\pm$6.84} & 56.8 \scriptsize{$\pm$12.0} & 36.9 \scriptsize{$\pm$7.32} & \textbf{60.0} \scriptsize{$\pm$10.6}& 58.7& 51.0 & \textbf{55.6}  \\
        
        &RTGNN~\cite{qian2023robust}  &  69.7 \scriptsize{$\pm$1.76} & 65.1 \scriptsize{$\pm$1.2} & 51.7 \scriptsize{$\pm$0.18} & 36.4 \scriptsize{$\pm$0.47} & 62.9 \scriptsize{$\pm$0.66} & 50.5 \scriptsize{$\pm$1.3} & 32.8 \scriptsize{$\pm$7.5} & 34.7 \scriptsize{$\pm$9.15} & 22.9 \scriptsize{$\pm$9.69} & 34.4 \scriptsize{$\pm$6.95}& 56.1& 31.2 & 46.1  \\
        
        &CLNode~\cite{wei2023clnode}  & 73.0 \scriptsize{$\pm$2.08} & \textbf{\underline{67.5}} \scriptsize{$\pm$0.85} & 66.1 \scriptsize{$\pm$0.29} &\textbf{ 40.2} \scriptsize{$\pm$0.36} & 65.9 \scriptsize{$\pm$1.04} &\textbf{\underline{64.9}} \scriptsize{$\pm$0.33} & 34.7 \scriptsize{$\pm$8.22} & 34.9 \scriptsize{$\pm$9.55} & 29.3 \scriptsize{$\pm$5.69} & 35.8 \scriptsize{$\pm$3.77}& \textbf{\underline{62.9}}& 33.7 & 51.2  \\
        
        &CGNN~\cite{yuan2023learning} & 73.6 \scriptsize{$\pm$2.96} & 60.0 \scriptsize{$\pm$12.4} & 66.8 \scriptsize{$\pm$0.41} & 31.4 \scriptsize{$\pm$6.61} & 60.9 \scriptsize{$\pm$2.99} & 40.6 \scriptsize{$\pm$18.9} & 25.0 \scriptsize{$\pm$10.5} & 34.1 \scriptsize{$\pm$6.53} & 29.1 \scriptsize{$\pm$6.00} & 35.1 \scriptsize{$\pm$7.07}& 55.5& 30.8 & 45.7  \\

        &CRGNN~\cite{li2024contrastive} & \textbf{74.7} \scriptsize{$\pm$1.45} & 64.3 \scriptsize{$\pm$1.39} & 5.6 \scriptsize{$\pm$0.00} & 34.0 \scriptsize{$\pm$1.73} & 62.3 \scriptsize{$\pm$0.84} & 54.6 \scriptsize{$\pm$4.38} & 21.1 \scriptsize{$\pm$5.88} & 37.3 \scriptsize{$\pm$10.2} & 16.4 \scriptsize{$\pm$10.1} & 26.2 \scriptsize{$\pm$6.88}& 49.3& 25.3 & 39.7  \\

        &PIGNN~\cite{du2021noise} &73.3 \scriptsize{$\pm$1.85} & 65.2 \scriptsize{$\pm$0.12} &OOM  & OOM & OOM & OOM & 23.7 \scriptsize{$\pm$10.6} & 33.0 \scriptsize{$\pm$10.2} & 28.9 \scriptsize{$\pm$7.03} & 32.3 \scriptsize{$\pm$7.18}& N.A.& 29.5 & N.A.  \\
        
        &RNCGLN~\cite{zhu2024robust} &  66.5 \scriptsize{$\pm$1.72} & 56.0 \scriptsize{$\pm$1.48} & OOM  & OOM & OOM & OOM & \textbf{57.4} \scriptsize{$\pm$14.2} & \textbf{\underline{63.8}} \scriptsize{$\pm$8.9} & \textbf{\underline{60.4}} \scriptsize{$\pm$5.24} & \textbf{\underline{62.1}} \scriptsize{$\pm$5.02}& N.A.& \textbf{\underline{60.9}} & N.A.  \\

        &R2LP~\cite{cheng2024resurrecting} & 72.9 \scriptsize{$\pm$1.27} & 20.4 \scriptsize{$\pm$11.7} & OOM  & OOM & OOM & OOM & 49.2 \scriptsize{$\pm$7.9} & 58.4 \scriptsize{$\pm$11.4} & 38.9 \scriptsize{$\pm$8.09} & 40.2 \scriptsize{$\pm$2.54}& N.A.& 46.7 & N.A.  \\
        \bottomrule
    \end{tabular}
    }
\end{table*}

Table~\ref{tab:detection} presents the performance of four supervised detection models with varying backbone architecture.
While GraphSAGE performs well, consistent with the results in Section~\ref{sec:label_noise_impact}, MLP achieves the highest overall performance, highlighting the importance of leveraging node features in detection tasks. 
Specifically, MLP excels when node features contain richer contextual information (e.g., Children and History) beyond simple bag-of-words representations (e.g., Cora-ML), as well as in handling Heterophilic graphs, where individual node features play a more important role.

\begin{figure}[t]
	\centering
	\begin{subfigure}{0.49\columnwidth}
	    \centering
	    \includegraphics[width=\textwidth]{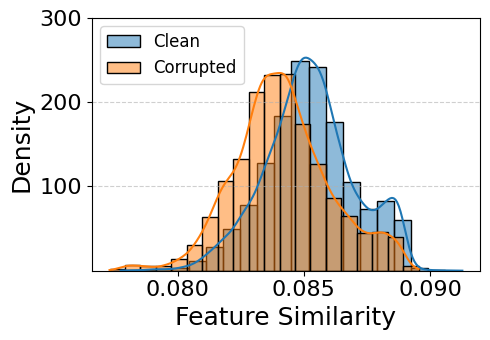}
	    \caption{Feature similarity}
        \label{fig:feat_similarity}
	\end{subfigure}
	\hfill
    \begin{subfigure}{0.47\columnwidth}
	    \centering
	    \includegraphics[width=\textwidth]{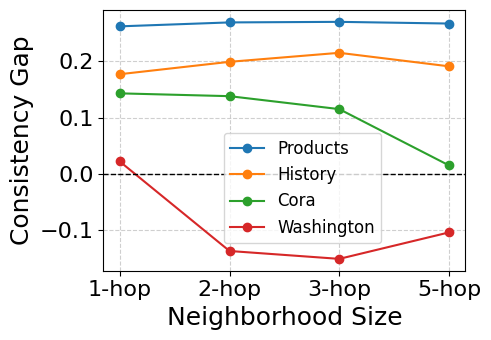}
	    \caption{Topological consistency gap}
        \label{fig:topo_consistency}
	\end{subfigure}

\caption{Feature similarity distribution and topological label consistency gap of the Photo dataset.}
\label{fig:graph_data_sim}
    \vspace{-0.2cm}
\end{figure}

\subsubsection{\textbf{Applicability of other information for noisy label detection}}   
\label{sec:detector_feattopo}
We further explore the applicability of two types of information: feature similarity and topology-based label consistency.

\smallsection{Feature similarity.}
We investigate whether clean and corrupted instances exhibit distinct characteristics from a feature perspective.
To this end, we measure the similarity between input features and their corresponding class representations for clean and corrupted instances, respectively.\footnote{The similarity is computed in the same way as the feature-based noise in Section \ref{sec:noise_modeling}}
Figure~\ref{fig:feat_similarity} compares their distributions. 
While the distributions of clean and corrupted data are somewhat distinguishable, the difference is not significant, especially compared to image data, where feature similarity-based models~\cite{zhu2022detecting} achieve competitive detection performance by leveraging rich contextual features from pre-trained models~\cite{radford2021learning}.
This suggests that solely relying on node features can be insufficient for graph data, particularly when node features are less informative. 
It also highlights the necessity of incorporating auxiliary features to supplement the lack of information. 
Exploring such multimodal approaches offers promising directions for future work.

\smallsection{Topology-based label consistency.}
We analyze whether the label consistency between a target node and its neighbors correlates with label corruption.  
To this end, we introduce the topology-based label consistency score and examine the score gap between clean and corrupted data.
The consistency score of a node $v_i$ for $k$-hop neighbors is defined as
$S_k(v_i) =\frac{1}{|\mathcal{N}_{k}(v_i)|} \sum_{v_j \in \mathcal{N}_{k}(v_i)} \mathbb{I}(y_i = y_j)$, where $\mathcal{N}_{k}(v_i)$ denotes $k$-hop neighbors of node $v_i$.
We then compute the average score gap between the clean and corrupted instances: $\Delta S_k=\frac{1}{|V^c|} \sum_{v_i \in V^c} S_k(v_i) - \frac{1}{|V^n|} \sum_{v_i \in V^n} S_k(v_i)$, where $V^c$ and $V^n$ represents the sets of clean and corrupted instances, respectively.

Figure \ref{fig:topo_consistency} presents the consistency gap across varying neighborhood sizes across four datasets. 
We observe a gap in consistency scores between clean and corrupted instances, with the impact of neighborhood size differing across datasets.
Notably, the tendency varies significantly between homophilic and Heterophilic graphs;
In the Heterophilic graph (i.e., Washington), the consistency gap is negative, indicating that corrupted instances exhibit higher label consistency with their neighbors than clean instances. 
This aligns with the nature of Heterophilic graphs, where neighboring nodes often have different labels. 
While topology-based label consistency can serve as an indicator for noisy label detection, its application to Heterophilic graphs may require adopting distinct—or even opposite—strategies compared to those used for homophilic graphs.

\subsection{Noise Robust Learning in Graph Data}
\label{sec:robust_learning}

We explore the effectiveness of various strategies for noise robust learning.
Our experiments evaluate 23 different methods, categorized into three types:
(1) \textbf{Base models} trained with the standard classification learning.
(2) \textbf{Learning with Label Noise (LLN)}, which introduces noise-robust learning techniques applicable to various base models.
(3) \textbf{GNN-specific LLN (G-LLN)}, which introduces LLN techniques tailored for Graph data.
GCN, the most widely used baseline model for node classification tasks, is used as a base model for both LLN and G-LLN.

\smallsection{Hyper-parameters.}
To ensure a fair comparison, we apply the same hyperparameter tuning process across all baseline models.
We utilize the Weights \& Biases (wandb) package\footnote{https://wandb.ai/site/models/} with random search for optimization and determine the optimal hyperparameters for each dataset-model combination (10 datasets × 23 models) under LLM label noise.
This allows for an efficient and systematic exploration of the search space, ensuring that each model receives well-tuned parameters.
For training, each model is generally run for a maximum of 200 epochs. However, for models requiring extended learning due to complexity or convergence challenges, we increase the training limit to 500 epochs.
Table~\ref{tab:hyper_param} in the Appendix presents the hyperparameter search space, including both general and model-specific configurations.
Most of the compared methods were built upon the PyTorch implementation by the previous work~\cite{wang2024noisygl}.


\smallsection{Performance of noise-robust models.}
Table \ref{tab:robust_exp} presents classification accuracy for each method, along with the average accuracy for the homophily, heterophily, and overall datasets.
As discussed in Section \ref{sec:label_noise_impact}, MLP, which relies solely on node features, shows competitive performance in Heterophilic graphs where label agreement among neighboring nodes is low.
Among various methods, CLNode, which assesses instance difficulty and employs curriculum learning, performs best on homophilic datasets, while RNCGLN, which integrates contrastive loss, self-attention, and self-improvement, excels in heterophily settings. 
However, most LLN and G-LLN models struggle to perform well across both cases.
Meanwhile, although GraphSAGE does not rank highest in any specific setting, it consistently delivers strong average performance. 
This suggests that GraphSAGE could serve as a reliable starting point for developing robust models to handle label noise.

\begin{figure}[t]
    \centering
    \includegraphics[scale=0.45]{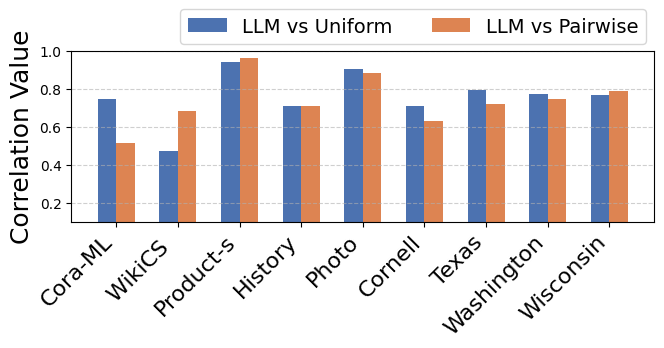}
    \caption{
    Correlation between classification accuracy on LLM-based noise and two types of class-dependent noise.}
    \label{fig:correl_other}
    \vspace{-0.3cm}
\end{figure}

\smallsection{Comparing class-dependent and LLM-based label noise.}
We investigate whether models that are robust to class-dependent noise also perform well against LLM-based noise.
Figure~\ref{fig:correl_other} presents the correlation between the performance of GraphSage on uniform/pairwise noise and LLM-based noise. 
We observe that the correlation varies across datasets. 
Notably, some datasets, such as Cora-ML and WikiCS, exhibit more pronounced deviations, indicating that LLM-based noise behaves uniquely in certain cases. 
This highlights the need for models specifically designed to handle more challenging and realistic noise patterns simulated by LLMs.

\begin{figure}[tbp]
	\centering
    \resizebox{0.9\columnwidth}{!}{ 
	\begin{subfigure}{0.45\columnwidth}
	    \centering
	    \includegraphics[width=\textwidth]{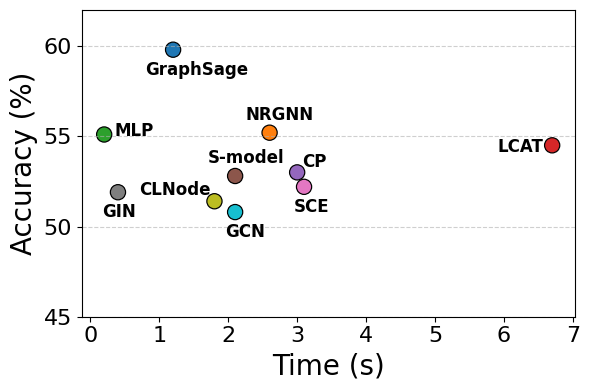}
	    \caption{Time consumption}
	\end{subfigure}
	\hfill
    \begin{subfigure}{0.45\columnwidth}
	    \centering
	    \includegraphics[width=\textwidth]{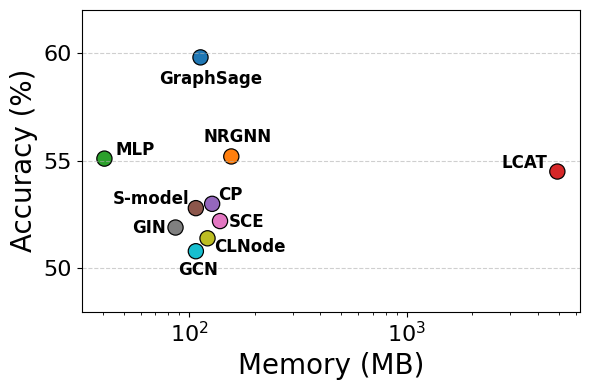}
	    \caption{Peak GPU memory}
	\end{subfigure}
    }
\caption{Time and memory usage comparison on the WikiCS dataset for the top 10 models ranked by average accuracy.}
\label{fig:memory}
    \vspace{-0.5cm}
\end{figure}

\smallsection{Training efficiency.}
Figure~\ref{fig:memory} shows the time and memory usage of the top 10 models (ranked by average accuracy) on the WikiCS dataset, evaluated using an NVIDIA GeForce RTX 3090 (24GB). Similar trends are observed across other datasets.

\section{Related Work}
Throughout Sections~\ref{sec:Label Noise for Graph Data} to~\ref{sec:Label Noise handle for Graph Data}, we present a detailed discussion of prior efforts to simulate noise and develop learning strategies for noisy data.
Here, we provide a concise summary of relevant~works.

\smallsection{Learning with label noise on graph data.}
While many LLN strategies have been developed for image data~\cite{goldberger2017training, patrini2017making, han2018co, wang2019symmetric, wei2020combating, ma2020normalized, li2020dividemix}, they are not always applicable to graph learning tasks due to the unique characteristics of graph data and GNNs~\cite{li2021unified}.
Specifically, many existing LLN methods often assume that data points are independent and identically distributed, which is not the case with graph-structured data, where nodes are interconnected and interdependent~\cite{dai2021nrgnn}. 
Furthermore, GNNs' message-passing mechanism adds additional complexity, as neighboring nodes influence each other’s features and labels, leading to noise propagation~\cite{zhu2024robust, dai2021nrgnn}. 

To address this challenge, researchers have developed various graph-specific LLN methods for robust learning, which can be categorized based on their core strategies for handling noise: 
loss regulation techniques, such as loss correction~\cite{hoang2019learning}, which adjust the loss to mitigate the impact of noisy labels;
robust training strategies, which improve model training by distinguishing between trustworthy and untrustworthy nodes~\cite{wei2023clnode};
graph structure augmentation, which refines or enhances the graph structure to reduce noise propagation~\cite{dai2021nrgnn, qian2023robust}; and contrastive learning, which leverages contrastive objectives to learn more robust node representations~\cite{yuan2023learning, li2024contrastive}.

However, no systematic study has been conducted on detecting label noise in graph datasets, and existing methods are limited to evaluating only naive class-dependent label noise, making it difficult to assess their effectiveness in more realistic scenarios.

\smallsection{Benchmarking and analyzing the impact of label noise.}
In other domains, such as computer vision, dedicated efforts have been made to benchmark and analyze the impact of label noise. 
These efforts have driven the evolution of more realistic noise simulation techniques and the development of robust learning methods.
Research has progressed from naive uniform or pairwise noise~\cite{van2015learning, ghosh2017robust, yu2023delving} to instance-dependent noise~\cite{xia2020part, jiang2020beyond}, followed by the introduction of various human-annotated noisy datasets~\cite{peterson2019human, xiao2015learning, wei2021learning}, which have facilitated significant advancements in model development.
However, to the best of our knowledge, only uniform and pairwise noise has been explored in graph data~\cite{hoang2019learning,  dai2021nrgnn, qian2023robust,wei2023clnode, du2021noise, li2024contrastive,  cheng2024resurrecting,  wang2024noisygl}, highlighting the need for more realistic noisy datasets explicitly designed for graphs, as well as a comprehensive benchmark.

\section{Conclusion and Future Study}
We propose BeGIN, a benchmark for label noise in graph data, offering real-world datasets with various noise types and a comprehensive evaluation of noise-handling strategies.
Through our analyses, we examine the impact of different noise types on GNNs and evaluate the effectiveness of various noise-handling strategies.
We also provide three key insights for future research: (1) structured mislabeling, particularly from LLM-based noise, poses significant challenges, 
(2) prioritizing central node features over naive neighbor aggregation improves model stability, 
(3) enhancing node features using pre-trained models or multi-modal information can further improve noise robustness.
We expect that BeGIN will be a valuable resource for driving research on label noise in graphs and promoting the development of robust GNN training approaches.

\smallsection{Ethics Statement.}
The proposed benchmark is built upon widely recognized, publicly available datasets for research purposes.
We utilize LLMs solely to generate noisy labels by selecting from the predefined categories within the dataset itself.
To ensure reproducibility and the integrity of our benchmark, we carefully document the LLM version, prompt structures, and noise generation process.
Our methodologies and findings do not cause harm to any individuals or groups, and we do not anticipate any significant ethical concerns arising from our work.

\begin{acks}
This work was supported by the NRF grant funded by the MSIT (No. RS-2024-00335873, RS-2023-00217286 
), and the IITP grant funded by the MSIT (No. 2018-0-00584, RS-2019-II191906 
).
This work was also supported by Basic Science Research Program through the NRF funded by the Ministry of Education (NRF-2021R1A6A1A03045425) and ICT Creative Consilience Program through the IITP grant funded by the MSIT (IITP-2025-RS-2020-II201819).
\end{acks}


\bibliographystyle{ACM-Reference-Format}
\bibliography{reference}

\appendix

\begin{table*}[h]
    \centering
    \caption{Dataset statistics and noise rate of LLM-based label noise.}
    \label{tab:A_dataset_stat}
        \resizebox{0.99\textwidth}{!}{%
        \begin{tabular}{ccccccc|cccc}
        \toprule
         &  \multicolumn{6}{c|}{Homophily datasets } & \multicolumn{4}{c}{Heterophily datasets }  \\

        &  Cora-ML  & WikiCS  & Product-s  & Children  & History  & Photo  & Cornell & Texas & Washington & Wisconsin \\
        \midrule

        \# Classes  & 7 & 10 & 44 & 24 & 12 & 12 & 5 & 5 & 5 & 5 \\
        
        \# Nodes  & 2,995 & 11,701 & 54,025 & 76,875 & 41,551 & 48,362 & 191 & 187 & 229 & 265 \\
        \# Edges  & 8,158 & 216,123 & 74,420 & 1,554,578 & 358,574 & 500,939 & 292 & 310 & 394 & 510  \\
         Avg. degree & 2.7&	18.5&	1.4&	20.2&	8.6&	10.4&	1.5&	1.7&	1.7&	1.9\\
        Node homophily  & 0.810 & 0.659 & 0.790 & 0.464 & 0.784 & 0.790 & 0.116 & 0.067 & 0.162 & 0.151  \\
        
        \# Node feat. & 2,879 & 300 & 100 & 768 & 768 & 768 & 1,703&1,703&1,703&1,703\\
        
        Node feat. range & Binary & -2 $\sim 2.5$ &  -220 $\sim 260$ &  -2 $\sim 13$ & -2 $ \sim 13$  & -2 $\sim 13$  &   Binary &  Binary &  Binary &  Binary \\ 

        Node  & Scientific paper &  Wikipedia page & Product & Book & Book & Product & \multicolumn{4}{c}{Web pages from CS departments} \\
        Edge  & Citation & Hyperlink &  \multicolumn{4}{c|}{Co-purcahased or co-views}  &  \multicolumn{4}{c}{Hyperlink} \\
        Text attr. & Bag-of-words & \multicolumn{4}{c}{Title \& contents} & User review & \multicolumn{4}{c}{Bag-of-words} \\
        Data source &  \cite{mccallum2000automating, bojchevski2017deep} & \cite{mernyei2020wiki} & \cite{hu2020open, he2023harnessing}  & \cite{ni2019justifying} &\cite{ni2019justifying} & \cite{ shchur2018pitfalls}& \cite{McCallum_4Universities} &\cite{McCallum_4Universities}&\cite{McCallum_4Universities}&\cite{McCallum_4Universities} \\

        Data processing & \cite{fey2019fast} & \cite{feng2024taglas} & \cite{ feng2024taglas} & \cite{yan2023comprehensive} & \cite{yan2023comprehensive}& \cite{yan2023comprehensive}& \cite{yan2023comprehensive}& \cite{yan2023comprehensive}& \cite{yan2023comprehensive}& \cite{yan2023comprehensive}\\

        Data split for robust learning & Random & Default & Default & Random & Random & Default & \multicolumn{4}{c}{Random}\\
        Split ratio (train/val/test) &  0.5/\ 0.1/\ 0.4  & 0.05/\ 0.15/\ 0.5 & 0.27/\ 0.03/\ 0.7 & 0.6/\ 0.2/\ 0.2 & 0.6/\ 0.2/\ 0.2 & 0.39/\ 0.15/\ 0.46 & \multicolumn{4}{c}{0.5/\ 0.3/\ 0.2}   \\

        \arrayrulecolor{gray} 
        \cmidrule(lr){1-11} 
        \arrayrulecolor{black}

        Noise rate (LLM-Naive) & 0.356 & 0.331 & 0.328 & 0.625 & 0.340 & 0.377 & 0.283 & 0.257 & 0.336 & 0.294 \\
        Noise rate (LLM-Reasoned) &0.403 & 0.383 & 0.385 & 0.617 & 0.399 & 0.394 & 0.314 & 0.278 & 0.345 & 0.309  \\
        Noise rate (LLM-Refined, default) & 0.306 & 0.309 & 0.303 & 0.575 & 0.322 & 0.356 & 0.272 & 0.246 & 0.314 & 0.283 \\
        
        \bottomrule
        
    \end{tabular}

    }

\end{table*}

\section{Additional Dataset Information}
\label{apx:dataset}
We present 10 benchmark datasets, all including text attributes on nodes. 
The details of these benchmarks are described in Table \ref{tab:A_dataset_stat}.

\smallsection{Cora-ML.}
The Cora-ML dataset comprises scientific papers categorized into seven groups. 
These papers form a citation network, ensuring that every paper in the dataset either cites or is cited by at least one other paper.
Each paper is represented using a bag-of-words (BoW) model~\cite{Harris01081954}.

\smallsection{WikiCS.}
WikiCS is a graph constructed from Wikipedia, where nodes represent page descriptions and edges correspond to hyperlinks between pages. 
The dataset, along with its raw text, is sourced from~\cite{mernyei2020wiki} and OFA~\cite{liu2023one}, and we utilize the processed version provided in~\cite{feng2024taglas}.
WikiCS is designed for node classification and includes 10 categories related to various computer science topics. The dataset provides multiple splits, and we adopt one for our experiments.

\smallsection{Products-s.}
The dataset is a subset of the co-purchasing network curated by TAPE~\cite{he2023harnessing} from the original OGB~\cite{hu2020open} dataset.
Nodes represent products on Amazon, and edges indicate that two products are bought together. 
Additional product descriptions serve as raw textual data. 
The objective is to classify products into one of several top-level categories in a multi-class classification task.

\smallsection{Children/ History.}
The dataset is a subset of the Amazon-Books dataset, consisting of books categorized under the second-level labels ``Children'' or ``History.''
Nodes represent books, and edges capture frequent co-purchases or co-views. Each book is assigned a three-level hierarchical classification label, with titles and descriptions serving as node text attributes. The objective is to classify books into categories related to children’s or historical literature.

\smallsection{Photo.}
The dataset is a subset of the Amazon-Electronics dataset, consisting of products categorized under the second-level label ``Photo.'' 
In this dataset, nodes represent electronics-related products, while edges capture frequent co-purchases or co-views between items. Each product is assigned a three-level hierarchical classification label. 
The text attribute for each product is derived from user reviews, where the most upvoted review is selected when available; otherwise, a random review is chosen. The objective is to classify products into various photography-related categories.

\smallsection{Cornell/ Texas/ Washington/ Wisconsin.}
The Cornell, Texas, Washington, and Wisconsin datasets are derived from a webpage dataset compiled by Carnegie Mellon University. 
They consist of web pages from the computer science departments of various universities, where nodes represent web pages, and edges correspond to hyperlinks between them.

\vspace{-0.1cm}

\section{Details of Label Corruption Process}
\label{apx:prompt}

\smallsection{Pairwise label noise.}
In Section~\ref{Sec:class_dependent}, we adhere to the standard practice of assigning labels to the subsequent class~\cite{wang2024noisygl}, which shifts a given class $i$ to the next class $(i~~\text{mod}~~C)+1$.

\smallsection{Confidence-based label noise.}
In Section~\ref{sec:noise_modeling}, we use the model prediction to estimate transition probabilities.   
To obtain these predictions, we perform five-fold data splitting (two folds for training, two for validation, and one for testing) and optimize the model until it achieves optimal validation performance. 
Prediction probabilities for the test set are stored for each fold. 
This procedure is repeated for 10 runs, and the results are averaged to compute the final transition probabilities.

\smallsection{LLM-based label noise.}
Section~\ref{sec:llm_noise} discusses the generation of LLM-based label noise.
We employ an LLM as a text classifier, providing explicit instructions to generate the most appropriate category along with a justification using GPT-4o-mini (OpenAI).
The prompt for Cora-ML is ``You are a domain expert in Machine Learning. Your task is to predict the most appropriate topic label for a research paper. You must choose from the following topic labels only:

To reduce sensitivity to variations in LLM responses, we refine the generated label noise by filtering out the simple case that can be correctly predicted without explicit reasoning. 
Specifically, we create LLM-based label noise twice: one (LLM-Naive) without requesting reasoning and the other (LLM-Reasoned) with requesting reasoning.
Then, we only adopt the LLM-reasoned labels in cases where both the LLM-naive and LLM-reasoned outputs disagree with the ground-truth label. 
This ensures that the retained noisy labels reflect more complex cases by filtering out simple misclassifications that could be corrected without deeper analysis.

\begin{itemize}[leftmargin=*]\vspace{-\topsep}
    \item \textbf{LLM-Naive}: Labels generated without requiring explicit reasoning or justification, relying solely on direct classification.
    \item \textbf{LLM-Reasoned}: Labels generated with explicit reasoning, providing a justification for the assigned category.
    \item \textbf{LLM-Refined (default)}: A refined noise set that prioritizes labels requiring explicit reasoning by filtering out simple misclassifications while integrating LLM-Naive and LLM-Reasoned.

\end{itemize}
The LLM-Refined has the lowest noise rate, whereas LLM-Reasoned exhibits the highest.
In this study, we adopt LLM-Refined as the default setting and maintain its noise rates consistently across all other noise types, such as uniform, pairwise, etc, unless otherwise specified.

\begin{figure}[tbp]
    \centering
    \includegraphics[scale=0.49]{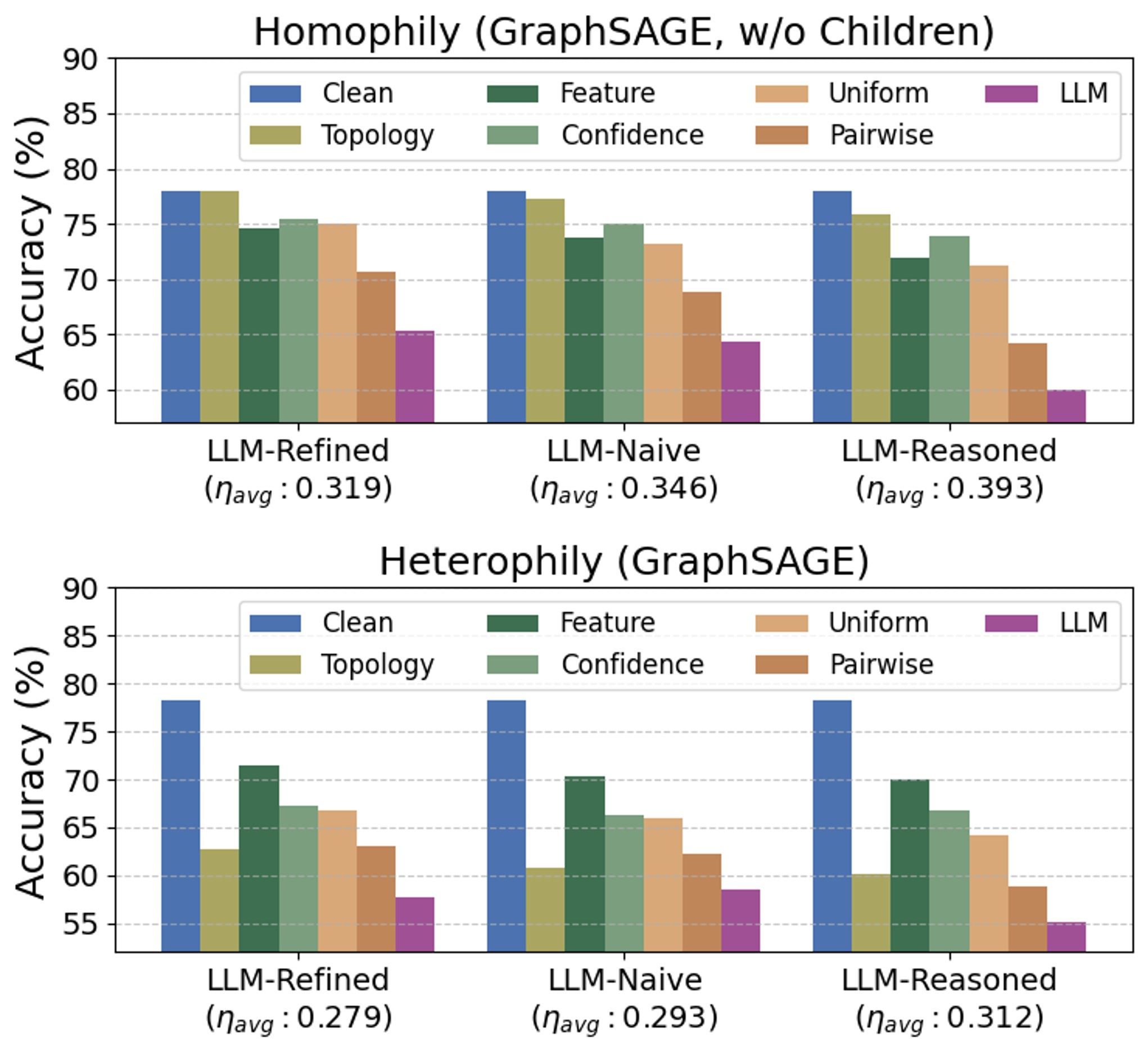}
    \caption{Average node classification accuracy on both homophily and heterophily datasets across different noise types and levels using GraphSAGE.}
    \label{fig:A_com_nois_gnn}
    \vspace{-0.4cm}
\end{figure}

\begin{table}[!b] 
    \centering
     \caption{Hyper-parameter search space.}
    \begin{adjustbox}{width=0.8\columnwidth} 
    \begin{tabular}{p{1.7cm}| p{6.3cm}} 
        \hline
        Model & Hyper-parameter search space \\
        \midrule
        SCE~\cite{wang2019symmetric} & $\alpha\in\{0.01, 0.1, 1.0 \}$\\\midrule
        JoCoR~\cite{wei2020combating} & $\lambda\in\{0.05, 
      0.1, 
      0.2,
      0.3,
      0.4,
      0.5,
      0.6,
      0.7,
      0.8,
      0.9,
      0.95 \}$\\\midrule
      \multirow{2}{*}{APL~\cite{ma2020normalized}} & $\alpha\in\{0.1, 
      1.0, 
      10.0, 
      100.0 \}$   \\
        & $\beta\in\{ 0.1, 
      1.0, 
      10.0, 
      100.0\}$  \\\midrule
      CP~\cite{zhang2020adversarial} & $\lambda\in\{0.05, 
      0.1,
      0.2,
      0.3 \}$\\\midrule
        
        \multirow{2}{*}{NRGNN~\cite{dai2021nrgnn}} & $\alpha\in\{0.01, 0.02, 0.03 \}$   \\
                & $\beta\in\{0.01, 0.1, 1, 10\}$  \\
        \midrule

        \multirow{2}{*}{RTGNN~\cite{qian2023robust}} & $\gamma\in\{ 0,
      0.05,
      0.1,
      0.2 \}$   \\
                & $\alpha\in\{0.03,
      0.1,
      0.3,
      1\}$  \\
                & $\lambda\in\{ 0.05,
      0.1,
      0.2\}$  \\
        \midrule

     \multirow{2}{*}{CLNode~\cite{wei2023clnode}} & $\lambda_0\in\{ 0.6,
      0.7,
      0.8,
      0.9,
      0.95 \}$   \\
                & $T\in\{ 50,
      100,
      150\}$  \\
        \midrule
        
          \multirow{2}{*}{CGNN~\cite{yuan2023learning} } & $\gamma\in\{0.25,
      0.5,
      0.75 \}$   \\
                & $\omega\in\{  0.6,
      0.7,
      0.8,
      0.9,
      0.95\}$  \\
        \midrule

                  \multirow{2}{*}{CRGNN~\cite{li2024contrastive}} & $\alpha\in\{  0.1,
      0.2,
      0.3,
      0.4,
      0.5,
      0.6,
      0.7,
      0.8,
      0.9,
      1 \}$   \\
                & $\beta\in\{  0.1,
      0.2,
      0.3,
      0.4,
      0.5,
      0.6,
      0.7,
      0.8,
      0.9,
      1\}$  \\
        \midrule
        RNCGLN~\cite{zhu2024robust} & $\alpha\in\{ 0.0001,
      0.001,
      0.1,
      1,
      10,
      100,
      1000 \}$\\\midrule

           \multirow{7}{*}{R2LP~\cite{cheng2024resurrecting}} & $\beta\in\{ 0.1,
      1,
      10,
      100,
      1000\}$   \\
                      & $\gamma\in\{ 0.1, 0.9\}$  \\

                       & $\alpha1\in\{ 0.1, 1.0\}$  \\
                       & $\alpha2\in\{ 0.1, 1.0\}$  \\
                       & $\alpha3\in\{ 0.1, 1.0\}$  \\
                       &Weight decay $\in$\\
         &$\{1e^{-7}, 5e^{-6},1e^{-6},5e^{-5},1e^{-5},5e^{-4}\}$\\
        \midrule

        \multirow{7}{*}{General} & Dropout $\in\{0.1,0.3,0.5, 0.7,0.9\}$\\
        & Number of hidden layer$\in\{1, 2,3,4\}$\\
        & Dimension of hidden layer $\in\{16, 32,64,128\}$\\
        & Learning rate $\in$\\
        &$\{1e^{-5},5e^{-5}, 1e^{-4},5e^{-4}, 1e^{-3}, 5e^{-3}, 1e^{-2}, 5e^{-2}, 1e^{-1}\}$\\
         &Weight decay $\in$\\
         &$\{1e^{-5}, 5e^{-5},1e^{-4},5e^{-4},1e^{-3},5e^{-3}, 1e^{-2}, 5e^{-2},1e^{-1}\}$\\

        \hline
    \end{tabular}
    \end{adjustbox}
   
    \label{tab:hyper_param}
\end{table}

\section{Additional Results for Robust Learning}
\label{apx:exp_robust}

\smallsection{Impact of varying noise levels}
We evaluate the classification accuracy on LLM-Naive and LLM-Reasoned to analyze the impact of varying noise levels on model performance. 
Figure~\ref{fig:A_com_nois_gnn} presents the summarized results for GraphSAGE.
Performance deteriorates as the noise rate increases, yet the overall trend remains consistent across different noise types, as discussed in Section~\ref{sec:label_noise_impact}.

\end{document}